%% file: manuscript.tex
\documentclass[conference]{IEEEtran}

\usepackage{cite}
\usepackage{amsmath,amssymb,amsfonts,amsthm}
\usepackage{graphicx}
\usepackage{textcomp}
\usepackage{xcolor}
\usepackage{tikz}
\usepackage{xurl} 
\usepackage{hyperref}
\usepackage{booktabs}
\usepackage{multirow}
\usepackage{lipsum}
\usepackage[linesnumbered, ruled, noline]{algorithm2e}

\ifCLASSOPTIONcompsoc 
\usepackage[caption=false,font=normalsize,labelfont=sf,textfont=sf]{subfig}
\else
\usepackage[caption=false,font=footnotesize]{subfig}
\fi
\newtheorem{definition}{Definition}
\def\BibTeX{{\rm B\kern-.05em{\sc i\kern-.025em b}\kern-.08em
    T\kern-.1667em\lower.7ex\hbox{E}\kern-.125emX}}

\begin{document}

\title{On Vessel Location Forecasting and the Effect of Federated Learning}

\makeatletter
\newcommand{\linebreakand}{%
  \end{@IEEEauthorhalign}
  \hfill\mbox{}\par
  \mbox{}\hfill\begin{@IEEEauthorhalign}
}
\makeatother

\author{\IEEEauthorblockN{Andreas Tritsarolis}
\IEEEauthorblockA{\textit{Department of Informatics} \\
\textit{University of Piraeus}\\
Piraeus, Greece \\
andrewt@unipi.gr}
\and
\IEEEauthorblockN{Nikos Pelekis}
\IEEEauthorblockA{\textit{Department of Statistics \&\ Insurance Science} \\
\textit{University of Piraeus}\\
Piraeus, Greece \\
npelekis@unipi.gr}
\and
\IEEEauthorblockN{Konstantina Bereta}
\IEEEauthorblockA{
\textit{Kpler}\\
Athens, Greece \\
kbereta@kpler.com}
\linebreakand %
\IEEEauthorblockN{Dimitris Zissis}
\IEEEauthorblockA{\textit{Department of Product \& Systems Design Engineering} \\
\textit{University of the Aegean}\\
Syros, Greece \\
dzissis@aegean.gr}
\and
\IEEEauthorblockN{Yannis Theodoridis}
\IEEEauthorblockA{\textit{Department of Informatics} \\
\textit{University of Piraeus}\\
Piraeus, Greece \\
ytheod@unipi.gr}
\and
}

\newcommand\copyrighttext{%
    \footnotesize To appear in the Proceedings of the 25th IEEE International Conference on Mobile Data Management (MDM), June 24 - June 27, 2024, Brussels, Belgium. \textcopyright 2024 IEEE. Personal use of this material is permitted. Permission from IEEE must be obtained for all other uses, in any current or future media, including reprinting/republishing this material for advertising or promotional purposes, creating new collective works, for resale or redistribution to servers or lists, or reuse of any copyrighted component of this work in other works.}
\newcommand\copyrightnotice{%
    \begin{tikzpicture}[remember picture,overlay]
        \node[anchor=south,yshift=15pt] at (current page.south) {\fbox{\parbox{\dimexpr\textwidth-\fboxsep-\fboxrule\relax}{\copyrighttext}}};
    \end{tikzpicture}%
}

\maketitle

\copyrightnotice

\begin{abstract}
    The wide spread of Automatic Identification System (AIS) has motivated several maritime analytics operations. Vessel Location Forecasting (VLF) is one of the most critical operations for maritime awareness. However, accurate VLF is a challenging problem due to the complexity and dynamic nature of maritime traffic conditions. 
    Furthermore, as privacy concerns and restrictions have grown, training data has become increasingly fragmented, resulting in dispersed databases of several isolated data silos among different organizations, which in turn decreases the quality of learning models.
    In this paper, we propose an efficient VLF solution based on LSTM neural networks, in two variants, namely Nautilus and FedNautilus for the centralized and the federated learning approach, respectively. We also demonstrate the superiority of the centralized approach with respect to current state of the art and discuss the advantages and disadvantages of the federated against the centralized approach.
\end{abstract}


\section{Introduction}\label{sect:Introduction}
    Vessel Location Forecasting (VLF) is a critical task in the maritime domain because it can be used in several aspects of maritime mobility awareness, including, among others, fishing effort/pressure forecasting \cite{Tampakis2021i4sea}, traffic flow management \cite{DBLP:conf/mbdw/MandalisCKPT22}, future collision avoidance \cite{DBLP:conf/mdm/TritsarolisCPT22}, as well as co-movement pattern discovery \cite{DBLP:conf/edbt/TritsarolisCTP21}. Informally, given a look-ahead time interval $\Delta t_{\text{next}}$, the goal is to predict the future location of a moving vessel $s_j$ at $\Delta t_{\text{next}}$ time after current timestamp $t^{s_j}_i$. 
    Figure \ref{fig:Future_Location_Prediction} illustrates such an example, where given the vessels' current routes (black solid lines), VLF predicts their corresponding future locations (green points).
    
    Nevertheless, the vast spread of IoT-enabled devices, such as sensors, smartwatches, smartphones, and GPS trackers, has led to the production of vast amounts of data, including mobility data. The availability of this volume of data is crucial to the success of Machine Learning (ML) technologies, which can perform a variety of tasks that may sometimes exceed human performance \cite{DBLP:series/synthesis/2019YangLCKCY}.
    Nevertheless, the information generated by the edge devices inherently consists of sensitive data and is frequently dispersed among numerous entities. These characteristics present novel challenges regarding the effective storage, analysis, and extraction of valuable insights from such data \cite{DBLP:journals/access/ZissisCSV20,DBLP:conf/bigdataconf/SpiliopoulosCZB17}.

    Centralizing data to a certain location (e.g., data center) may become quite a cumbersome task because of the high storage/bandwidth costs (e.g., commercial maritime traffic systems monitor thousands of vessels per day, receiving several TBs of AIS information). In addition, sharing data entails several risks including disclosure of commercial information, trade secrets, and customer personal information\footnote{c.f., for instance, the General Data Protection Regulation (GDPR) in European Union: \url{https://ec.europa.eu/info/law/law-topic/data-protection_en}.}.
    Therefore, in some domains, collecting and sharing data may become quite difficult, if not outright impossible, thus forcing data owners to store them in isolated data silos. Alternatively, delegating the training process to the edge devices and/or data silos, so that each party can use an ML-based model using their own datasets, may impact the models' performance, with sub-optimal performance (e.g., under-fitting) or a biased target distribution (e.g., over-fitting), depending on the datasets' size and features' distribution, respectively. 

    In order to solve the aforementioned challenges and train an ML-based model that does not rely on collecting all data to a centralized storage, McMahan et al. \cite{DBLP:journals/corr/McMahanMRA16} and Konečný et al. \cite{DBLP:journals/corr/KonecnyMYRSB16} propose the Federated Learning (FL) paradigm, where a centralized model is trained on decentralized data. In particular, each edge device (or data silo, depending on the problem architecture) receives a seminal model from the server and proceeds to train it using its corresponding data. Afterwards, all updated models are uploaded to the server, where they are aggregated, thus producing a new model. Repeating the process for several cycles may eventually cause the global model to converge, producing an ML model that competes, in terms of quality, the (local) model that each party can learn on its own \cite{DBLP:series/synthesis/2019YangLCKCY}.
    
    Using FL, the decentralized nature of the data is maintained, as the edge devices / data silos collaboratively train an ML model, only sending weight updates (i.e., gradients) to the aggregation server. Because of that, every participant keeps control of its own data, as it essentially never ``leaves'' the device / silo, therefore making it harder for an adversary to extract sensitive information. 
    Distributing the training workload to multiple edge devices / silos, FL allows for potentially ``smarter'' models, lower inference latency, less overall power consumption, and by extension, lighter environmental impact \cite{DBLP:journals/corr/abs-2102-07627}.  

    \begin{figure}[!ht]
      \centering
      \includegraphics[width=0.9\columnwidth]{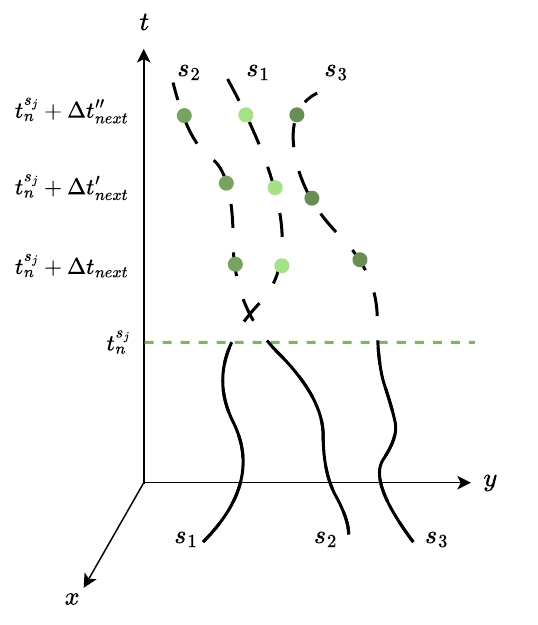}
      \caption{Predicting the future locations of three moving objects (e.g., vessels in the maritime domain) $s_1$, $s_2$ and $s_3$. Given the vessels' current routes up to $t^{s_j}_n$ (solid lines), VLF is used to predict vessels' future locations (green points) at three discrete timestamps, $t^{s_j}_n + \Delta t_{\text{next}}$, $t^{s_j}_n + \Delta t_{\text{next}}'$, and $t^{s_j}_n + \Delta t_{\text{next}}''$, respectively.}
      \label{fig:Future_Location_Prediction}
    \end{figure}
    
    In this paper, we propose Nautilus, an efficient VLF solution based on Neural Networks (NN), in particular the Long Short-Term Memory (LSTM) model, which will be shown (in Section \ref{sect:experiments}) to outperform current state-of-the-art \cite{DBLP:journals/tits/ChondrodimaPPT23} in the vast majority of cases at a prediction horizon $\Delta t_{\text{next}}$ up to 60 min. Moreover, with the FL paradigm in mind, we extend Nautilus to an FL-oriented architecture, called FedNautilus, geared towards collaboratively training the LSTM model across multiple data silos (i.e., vessel traffic controllers, fleet owners, maritime ICT industries, etc.). 

    The approach described in this paper meets the industrial needs of vessel tracking and maritime intelligence companies, like Kpler. Kpler owns the largest terrestrial network of AIS receivers worldwide as well as MarineTraffic\footnote{MarineTraffic: Global Ship Tracking Intelligence. \url{www.marinetraffic.com}}, one of the best known platforms for real-time vessel tracking, which is based on AIS. However, AIS is a collaborative maritime reporting system, so sometimes vessels might become untraceable via AIS, for the following reasons: (a) equipment malfunction, (b) the vessel navigates into an area that is out of AIS coverage, (c) the vessel has its AIS transponder switched-off (e.g., when engaging in illegal activities). VLF gives us the opportunity to estimate the future location(s) of a vessel given the last known positions. 

    The main motivation behind the FL approach is that since AIS data are collected using a de-centralized network of AIS receivers, scattered all over the world, the use of FL allows for reducing the data that needs to be transferred from the receivers to the centralized server, enabling the creation of local VLF models which are then aggregated into a global model in federated fashion. In addition, due to the diversity of the participating parties' datasets, the global model can also extract knowledge of certain behaviours (e.g., tight maneuvering) that may not be available to all parties. 
    Nevertheless, the FL problem in our setting is challenging since, apart from the inherent difficulty of forecasting, we also need to define the FL communication protocol (e.g., FedAvg \cite{DBLP:journals/corr/McMahanMRA16}), which is not a straightforward procedure at all.

    In summary, the main contributions of this work are as follows: 
    
    \begin{itemize}
        \item We propose an efficient VLF architecture based on LSTM, in two variants, namely Nautilus and FedNautilus, for the centralized and the (cross-silo) federated learning approach, respectively.
        \item We demonstrate the efficiency of the proposed architecture, in terms of prediction accuracy in short-term prediction horizon (up to 60 min.), using three large-volume real-world maritime AIS datasets.
        \item We study the effect of FL on the task at hand, i.e., the performance of FedNautilus with respect to FL aggregation hyper-parameters.
    \end{itemize}

    The rest of this paper is organized as follows: Section \ref{sec:RelatedWork} discusses related work. Section \ref{sec:Background_and_Definitions} formulates the problem at hand and presents the proposed (Fed)Nautilus architecture in its two variants (centralized vs. federated learning). Section \ref{sect:experiments} presents our experimental study, where it is shown that our solution outperforms state-of-the-art. 
    Finally, Section \ref{sec:conclusion} concludes the paper, also giving hints for future work.

\section{Related Work}\label{sec:RelatedWork}
    
    \subsection{Vessel Location Forecasting}    
        Considering the VLF problem, current state-of-the-art includes an adequate number of research works. More specifically, one line of work includes clustering-based prediction techniques. Petrou et al. \cite{DBLP:conf/ssd/PetrouNTGKSPVGC19} utilize the work done by \cite{DBLP:conf/bigdataconf/TampakisPDT19} on distributed subtrajectory clustering, in order to extract individual subtrajectory patterns from big mobility data; these patterns are subsequently utilized in order to predict the future location of the moving objects in parallel. In a more recent work, Zygouras et al. \cite{DBLP:journals/access/ZygourasTZ24} introduce EnvClus$^\ast$, a novel data-driven framework which performs trajectory forecasting via a mobility graph which models vessels' most likely movements among two ports.
                
        Wang et al. \cite{DBLP:journals/corr/abs-1906-04928} aiming at predicting the movement trend of vessels in the crowded port water of Tianjin port, proposed a vessel berthing trajectory prediction model based on bidirectional GRU (Bi-GRU) and cubic spline interpolation. Capobianco et.al. \cite{DBLP:journals/taes/CapobiancoMFBW21} provided an NN-based approach for vessel trajectory prediction. In particular, they predict the vessels' future locations using a temporal window within an area of interest, and an encoder-decoder LSTM using the attention mechanism. 

        Suo et al. \cite{DBLP:journals/sensors/SuoCCY20} present an RNN-based model to predict vessel trajectories based on the DBSCAN \cite{DBLP:conf/kdd/EsterKSX96} clustering algorithm to derive main trajectories, and a symmetric segmented-path distance approach to eliminate the influence of a large number of redundant data and optimize incoming trajectories. 
        
        Liu et al. \cite{DBLP:conf/icdm/Liu0SL19} propose ``Spatio-Temporal GRU'', a trajectory classifier for modeling spatio-temporal correlations and irregular temporal intervals prevalently presented in spatio-temporal trajectories. More specifically, a segmented convolutional weight mechanism was proposed to capture short-term local spatial correlations in trajectories along with an additional temporal gate to control the information flow related to the temporal interval information. 
        
        Most recently, Chondrodima et al. \cite{DBLP:journals/tits/ChondrodimaPPT23} propose a novel LSTM-based VLF framework, specially designed for handling vessel data by addressing some major GPS-related obstacles, such as variable sampling rate and sparse trajectories. Moreover, in order to improve the predictive power of VLF, they propose a novel trajectory data augmentation method based on the well-known Douglas-Peucker line simplification algorithm \cite{doi:10.3138/FM57-6770-U75U-7727}. 
        This work is considered, to the best of our knowledge, the state of the art in (short-term) VLF achieving an accuracy error around 2 km in 30 min. prediction horizon. 

    \subsection{Federated Learning}
        While distributed ML \cite{DBLP:conf/www/LiuCW17} can help us scale up the training process across multiple computational nodes, it can only be used on centralized data. On the other hand, FL trains centralized models using decentralized data \cite{DBLP:conf/aistats/McMahanMRHA17}, as such, FL algorithms are primarily geared towards data privacy. 
        
        Considering the characteristics of the data owners, we distinguish two major FL variants, namely, cross-silo and cross-device FL \cite{DBLP:journals/ftml/KairouzMABBBBCC21}. Cross-device FL can be considered when the participating devices (clients) are typically large in number (up to $10^{10}$) and have slow or unstable internet connection; a principal motivating example arises when the training data comes from users’ interaction with mobile applications \cite{DBLP:journals/corr/KonecnyMRR16}. On the other hand, cross-silo FL can be considered when a relatively small group (usually 2--100) of organizations share a common incentive to collaboratively train an ML model based on their data, but cannot share them directly, due to e.g, storage cost or legal constraints. Another key difference between cross-device and cross-silo FL lies within the privacy requirements of the FL framework. In cross-device FL, data privacy is of the utmost importance, as the trained ML model will be available to virtually everyone, whereas in cross-silo FL, the trained ML model most likely be available for internal use among the participating parties, therefore the concerns about ``virtually everyone'' are less important in the life cycle of the ML model.

        Except network and communication efficiency and client availability \cite{DBLP:journals/corr/KonecnyMYRSB16}, another key challenge of federated optimization is the training parties' heterogeneity with respect to their local datasets \cite{DBLP:journals/corr/KonecnyMRR16}. In order to address the first two issues, FedAvg \cite{DBLP:journals/corr/McMahanMRA16} performs multiple local updates on the available clients before communicating to the server. While this approach works well with high convergence guarantees (in applications where the participating parties' datasets are homogeneous), when the clients are heterogeneous these guarantees fail to hold. By each step, the parties' locally fitted ML model will converge to different local optima, therefore introducing slow and unstable convergence to the global model, as Figure \ref{fig:client-drift-example} illustrates. This phenomenon is known as ``client-drift'', and in order to avoid it, fewer local updates and/or lower learning rates must be used, actions that have a significant impact on the convergence stability of FedAvg.
            
        \begin{figure}
              \centering
              \includegraphics[width=.9\columnwidth]{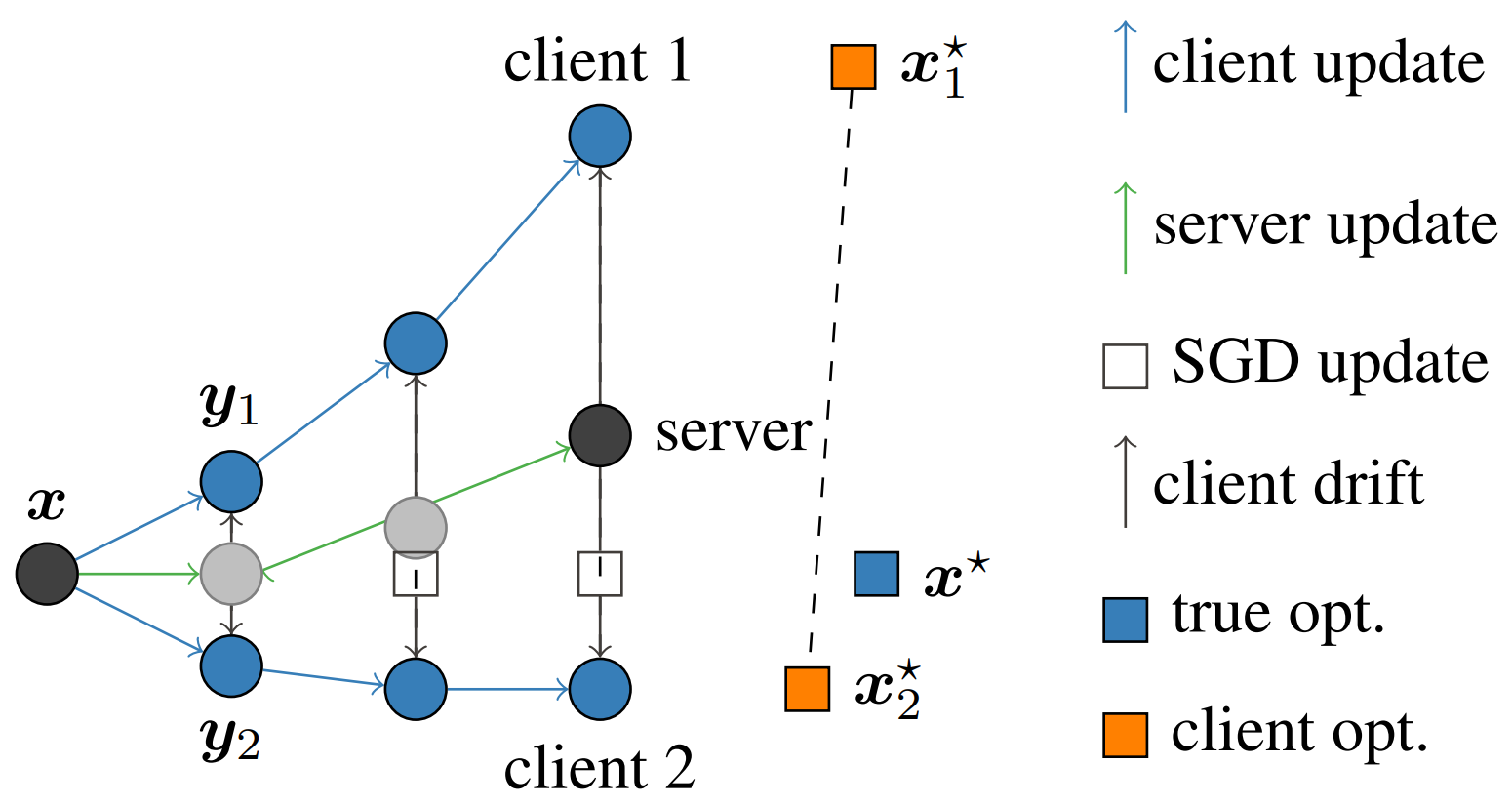}
              \caption{Client-drift in FedAvg is illustrated for 2 clients with 3 local steps ($N = 2$, $K = 3$). The local updates $y_i$ (in blue) move towards the individual client optima $x_i^\ast$ (orange square). The server updates (in green) move towards $\frac{1}{N} \sum_{i}{x_i^\ast}$ instead of to the true optimum $x^\ast$ (black square; \cite{DBLP:conf/icml/KarimireddyKMRS20}).}
              \label{fig:client-drift-example}
        \end{figure}
    
        \begin{figure*}[!t]
            \centering
            \includegraphics[width=\textwidth]{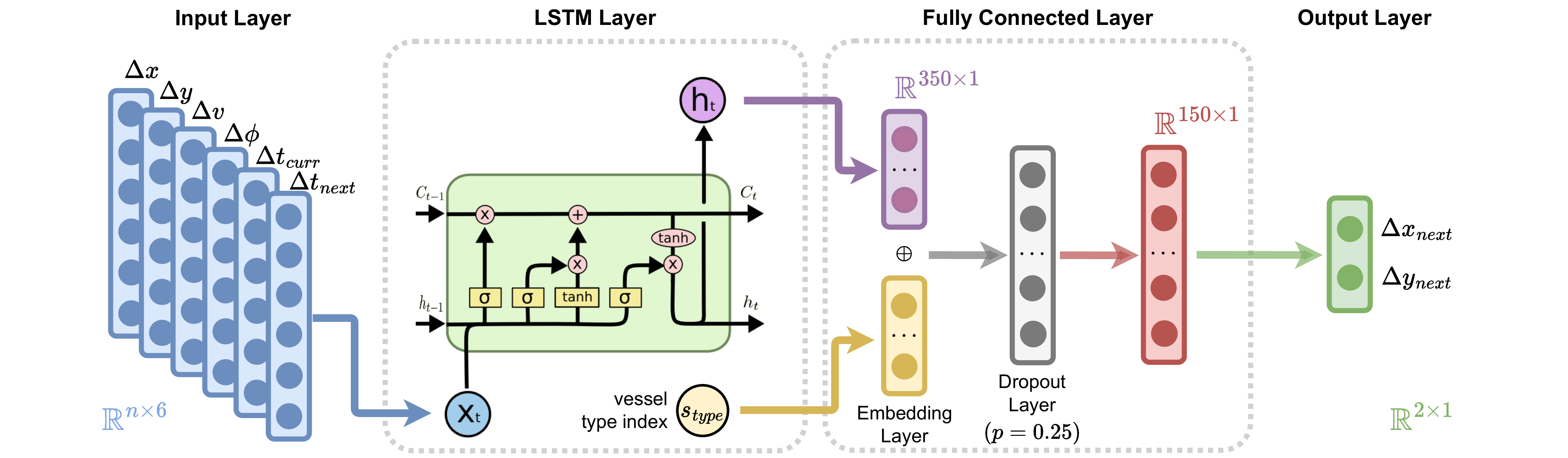}
            \caption{The proposed Nautilus architecture.}
            \label{fig:Nautilus}
        \end{figure*}
        
        Towards this direction, the authors in \cite{DBLP:conf/icml/KarimireddyKMRS20} acknowledge the aforementioned issue and propose a new federated optimization framework called SCAFFOLD, which uses control variates (variance reduction) in order to approximate an ideal unbiased update, therefore considering the ``client-drift'' in its local updates. By experimenting on various optimization settings, the authors prove that SCAFFOLD is resilient to client sampling (i.e., independent of the amount of client heterogeneity) and consistently outperforms FedAvg on non-convex experiments. Further following this line of research, FedProx \cite{DBLP:conf/mlsys/LiSZSTS20} presents an extension to FedAvg which adds a regularization term in the clients' cost function in order to restrict local updates to be closer to the initial (global) model. In a similar fashion, the qFedAvg algorithm \cite{DBLP:conf/iclr/LiSBS20} introduces a novel optimization objective inspired by fair resource allocation in wireless networks that encourages a more uniform accuracy distribution across devices in federated networks.    
        Closest to our work, however in the urban domain, the authors in \cite{DBLP:conf/cvpr/WangCW022} propose ATPFL, a framework for predicting pedestrian trajectories, combining FL with Automated Machine Learning (AutoML). Additionally, FAHEFL \cite{https://doi.org/10.1002/int.22987} is an FL-based algorithm geared towards user/model privacy (via Homomorphic Encryption) for vehicle trajectory prediction and behaviour classification.  
        
        To the best of our knowledge, our work is the first in the literature that experimentally evaluates centralized vs. federated VLF approaches over diverse (with respect to location and activity) maritime data silos, providing insightful results.

\section{Problem Formulation and Proposed Methodology}\label{sec:Background_and_Definitions}
    In this section, we present the proposed Nautilus solution and describe it under both centralized and federated learning approaches.

    \subsection{Problem Definition}
        Before we proceed to the actual formulation of the problem, we provide some preliminary definitions.
        
        \begin{definition}
            (Trajectory). A trajectory $T = \{p_1, \dots p_n\}$ of a moving object is defined as a sequence of timestamped locations, $p_i = \{x_i, y_i, t_i\}$, $1 \leq i \leq n$.
        \end{definition}
    
        \begin{definition}\label{def:VesselLocationForecasting}
            (Vessel Location Forecasting). Given a dataset $D$ of moving vessels' trajectories, a trajectory $T^s$ and the type $s_{type}$ of vessel $s$, and a prediction horizon $\Delta t_{next}$, the goal is to train a data-driven model over $D$, which will be able to predict the vessels' future location $p_{n+1}^s = \lbrace x_{n+1}^s, y_{n+1}^s, t_{n+1}^s \rbrace$ at timestamp $t_{n+1}^{s} = t_n^s + \Delta t_{next}$.
        \end{definition}
    
        If we recall Figure \ref{fig:Future_Location_Prediction}, it provides an illustration of Definition \ref{def:VesselLocationForecasting}. More specifically, we know the movement of three moving objects up to $t^{s_j}_{n}$. Our goal, given $\Delta t_{\text{next}}$, is to predict the anticipated location of these vessels at $t^{s_j}_{n} + \Delta t_{\text{next}}$.

    \subsection{The proposed Nautilus architecture}\label{subsect:Methodology}
        \input{tabs/dataset-stats} %
                        
        To address the VLF problem, we propose an extension of the LSTM-based model employed in \cite{DBLP:journals/tits/ChondrodimaPPT23} for predicting the future location of moving vessels in the short-term, which considers their kinematic characteristics and their corresponding type; Figure \ref{fig:Nautilus} illustrates the architecture of the proposed model. More specifically, Nautilus architecture consists of the following layers: a) an input layer of six neurons, one for each input variable, b) a single LSTM hidden layer composed of 350 neurons, c) an embedding layer with six dimensions for vectorizing the vessel's type, d) a Dropout layer with probability $p = 0.25$ for model regularization, e) a fully-connected hidden layer composed of 150 neurons, and f) an output layer of two neurons, one for each output variable. 

        The input variables consist of the differences (i.e., deltas) in the Cartesian space ($\Delta x, \Delta y$), speed $\Delta v$ and course $\Delta \phi$ over ground, current time $\Delta t_{\text{curr}}$, as well as the temporal horizon $\Delta t_{\text{next}}$ for which we want to predict the vessels' future location. On the other hand, the output variables consist of the differences in space ($\Delta x_{\text{next}}$, $\Delta y_{\text{next}}$) of the predicted locations with respect to the current locations.
        
        With respect to the employed LSTM, each cell includes three gates, namely, ``forget'', ``input'', and ``output'', which are responsible for what information we are going to drop from the cell state $\mathbf{c}_{t-1}$, store from the cell state $\mathbf{c}_{t-1}$, and output to the next cell state $\mathbf{c}_{t}$, respectively. Eqs. \ref{eq:ForgetGate}--\ref{eq:HiddenState} briefly state the update rules for the employed LSTM layer \cite{Hochreiter1997}. Additionally, details for the Back-Propagation Through Time (BPTT) algorithm, can be found in \cite{58337}.

        \begin{align}
                & \mathbf{f}_t = \sigma\left( \mathbf{W}_f x_t + \mathbf{R}_f h_{t-1} + \mathbf{b}_f \right) \label{eq:ForgetGate} \\
            %
                & \mathbf{i}_t = \sigma\left( \mathbf{W}_i x_t + \mathbf{R}_i h_{t-1} + \mathbf{b}_i \right) \label{eq:InputGate} \\
            %
                & \mathbf{o}_t = \sigma\left( \mathbf{W}_o x_t + \mathbf{R}_o h_{t-1} + \mathbf{b}_o \right) \label{eq:OutputGate} \\
            %
                & \mathbf{\tilde{c}}_t = \tanh{\left( \mathbf{W}_c x_t + \mathbf{R}_c h_{t-1} + \mathbf{b}_c \right)} \label{eq:CandidateCellState} \\
            %
                & \mathbf{c}_t = \mathbf{f}_t \odot \mathbf{c}_{t-1} + \mathbf{i}_t \odot \mathbf{\tilde{c}}_{t-1} \label{eq:CellState} \\
            %
                & \mathbf{h}_t = \mathbf{g}_t \odot \tanh{\left( \mathbf{c}_t \right)} \label{eq:HiddenState}
        \end{align}
    
        \noindent
        where $\mathbf{h}_t$ is the hidden state at time $t$, $\mathbf{c}_t$ ($\mathbf{\tilde{c}}_t$) is the (intermediate) cell state at time $t$, $x_t$ is the input at time $t$, $\mathbf{h}_{t-1}$ is the hidden state of the LSTM cell at time $t-1$ or the initial hidden state at time $0$, and $\mathbf{i}_t$, $\mathbf{f}_t$, $\mathbf{g}_t$, $\mathbf{o}_t$ are the input, forget, cell, and output gates, respectively. $\mathbf{W}$, $\mathbf{R}$ and $\mathbf{b}$ are the input weight matrices, the recurrent weight matrices and the bias terms, respectively. $\sigma$ is the sigmoid function and $\odot$ is the Hadamard product. 
                
        For the centralized training approach, we train our model using the Adam \cite{DBLP:journals/corr/KingmaB14} optimization algorithm with learning rate $\eta = 10^{-4}$ and the Root Mean Squared Error (RMSE) loss function for 100 epochs. To prevent over-fitting, we use the well-known early stopping \cite{DBLP:series/lncs/Prechelt12} mechanism with a patience of 10 epochs. 

        Our model extends \cite{DBLP:journals/tits/ChondrodimaPPT23} by dividing its architecture into two parts based on the type of mobility information: internal and external. The former is related to the input of the recurrent layer, which encodes vessels' trajectories, whereas the latter enriches the projection of the vessels' input trajectory at a fully connected layer. In terms of internal information, we include additional factors related to vessels' movement, such as $\Delta v$ and $\Delta \phi$. Furthermore, in terms of external information, we enrich the projection of the input trajectory with information related to the type of the vessel via a separate embedding layer that is jointly trained with the model.
        
    \subsection{Extending to FedNautilus}\label{subsect:FedVLF}
        \begin{figure*}[!ht]
            \centering
            \includegraphics[width=\textwidth]{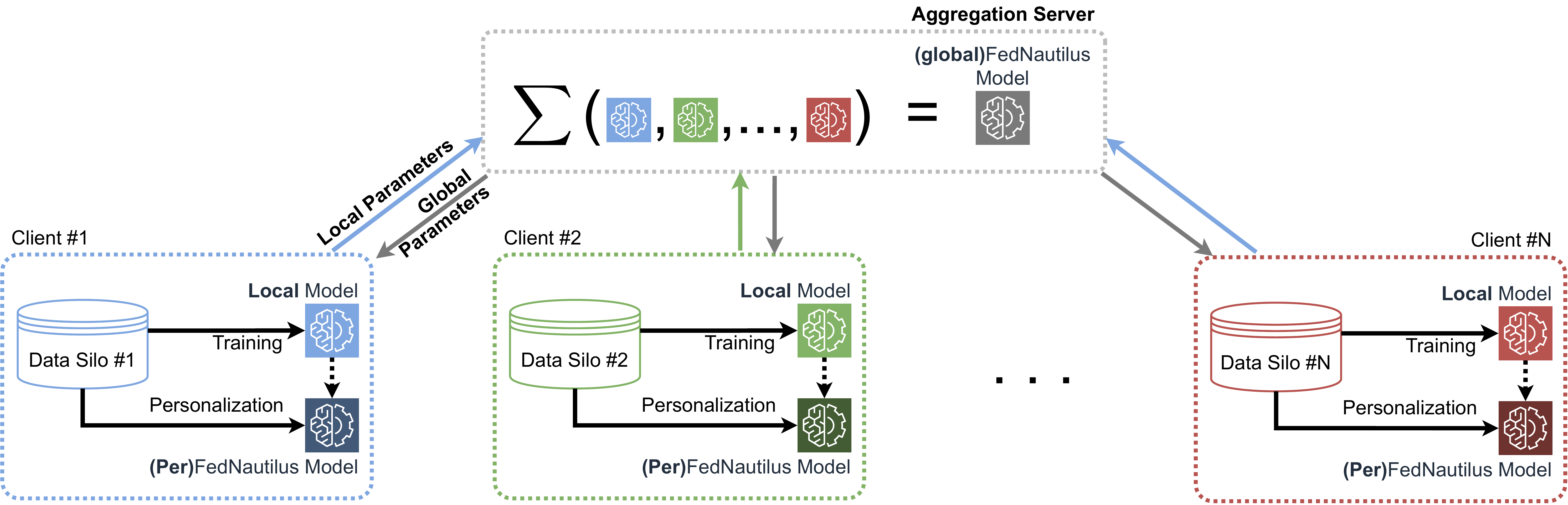}
            \caption{Extending to FedNautilus - Federated Learning Workflow}
            \label{fig:fednautilus-workflow}
        \end{figure*}

        Figure \ref{fig:fednautilus-workflow} illustrates the FL workflow of FedNautilus. In general, we have $N$ clients, each of which trains a (local)FedNautilus instance with their own data (stored in separate data silos). Afterwards, they share the parameters (i.e., weights) of their local models to the aggregation server, which generates the (global)FedNautilus model and returns it to the participating clients. Moreover, in order to tailor (local)FedNautilus' decisions to the clients' needs, we fine-tune the (local)FedNautilus model for a -- usually -- small number of epochs, a technique known as Personalized Federated Learning (PerFL) \cite{9743558}, which will be explained further in the following paragraphs.
        
        In our FL environment, each data silo corresponds to the transmitted locations of a certain AIS-enabled fleet. Each silo contains an instance of our Nautilus model, which is trained using only their respective data. For training the local model instances, we use the same optimization algorithm with the centralized approach, while for the optimization of the global model we use the FedProx \cite{DBLP:conf/mlsys/LiSZSTS20} algorithm for 70 FL rounds with $\mu_{prox} = 10^{-3}$, empirically selected based on the values listed at \cite{DBLP:conf/mlsys/LiSZSTS20} and overall training progress of the corresponding (local)Nautilus models.
        
        Furthermore, to further address the client-drift issue, we exploit on Personalized Federated Learning (PerFL) \cite{9743558}. PerFL is a branch of FL that aims to address these issues by customizing the global model for each client in the federation. In a nutshell, it aims to leverage the collective wisdom of clients' data in order to create models that are tailored to the data distributions of individual clients. In this paper, we follow the approach described in \cite{DBLP:conf/icc/TaikC20}, which is to fine-tune the global model for a certain number of epochs (10 in our case) using the clients' corresponding datasets, and extend it by using early stopping \cite{DBLP:series/lncs/Prechelt12} mechanism with a patience of 3 epochs, in order to avoid over-fitting.

\section{Experimental Study} \label{sect:experiments}
    In this section, we evaluate our VLF model on various centralized and federated learning schemes using three real-world maritime mobility datasets, and present our experimental results\footnote{The corresponding source code used in our experimental study is available at: \url{https://github.com/DataStories-UniPi/Nautilus}}.
    
    \subsection{Experimental Setup, Datasets and Preprocessing}\label{subsect:ExperimentalSetup}        
    
    \begin{figure*}
        \centering
        \subfloat[\label{subfig:BrestDataset}]{%
            \includegraphics[width=0.49\columnwidth]{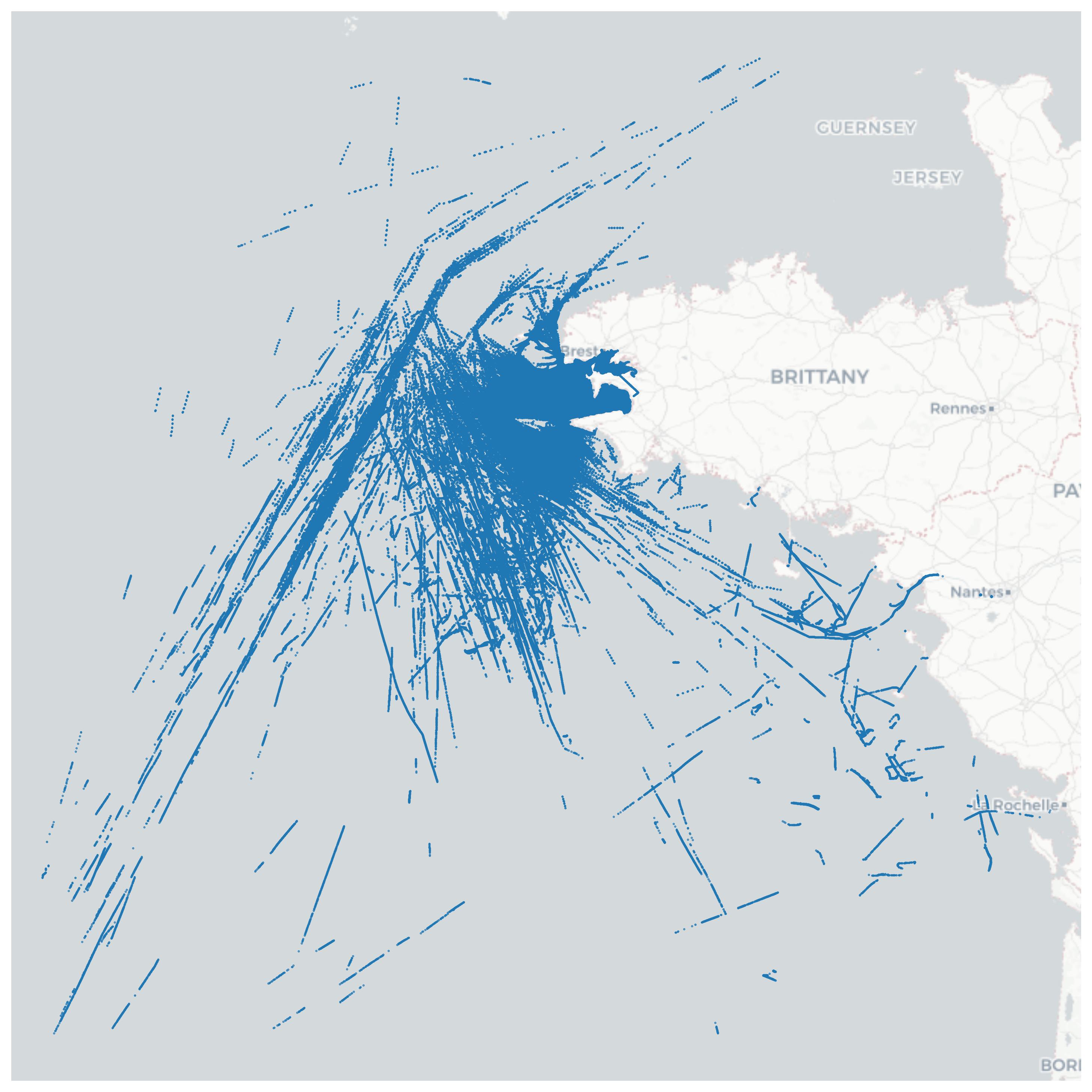}}
        \subfloat[\label{subfig:SaronicDataset}]{
            \includegraphics[width=0.49\columnwidth]{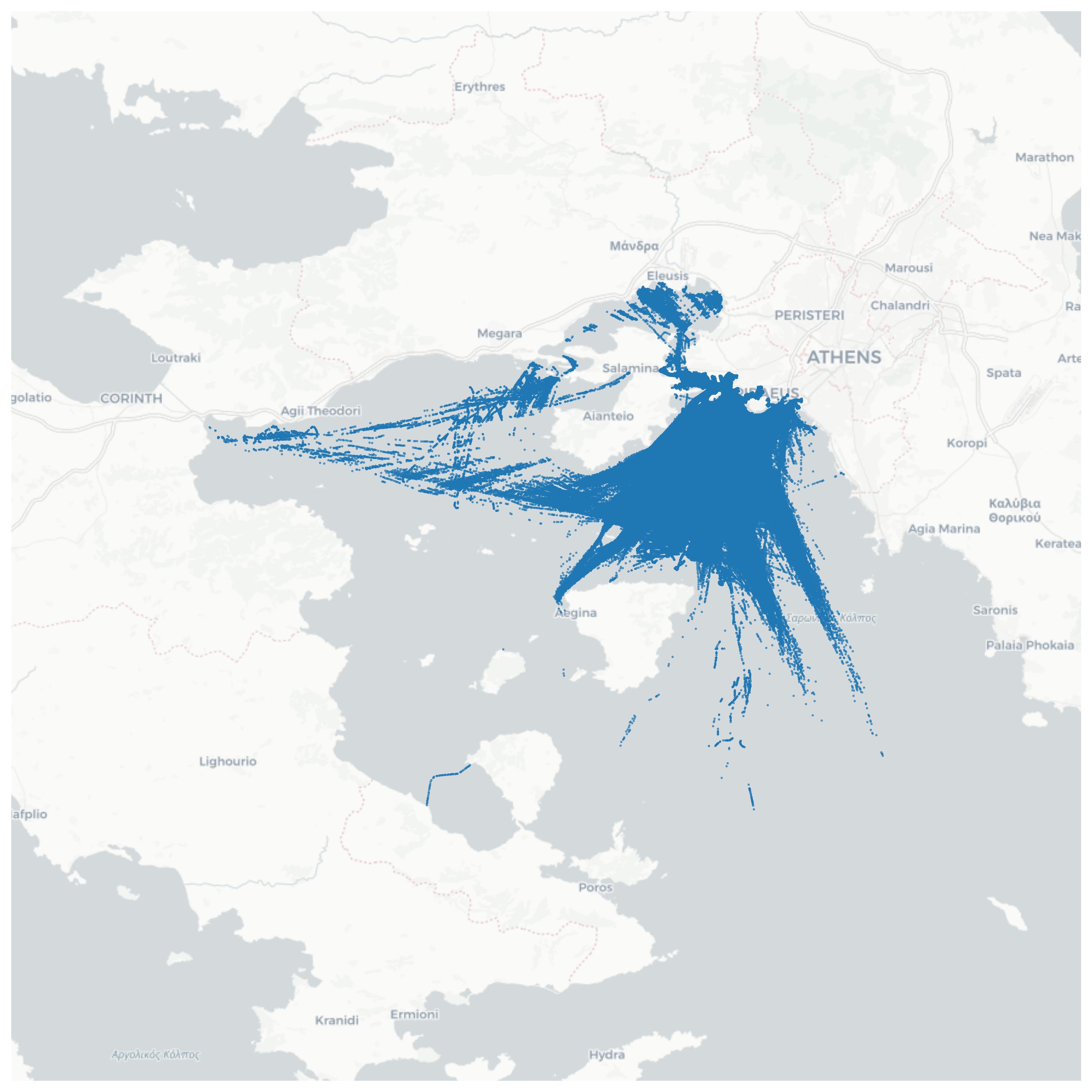}}
        \subfloat[\label{subfig:MarineTrafficDatast}]{%
            \includegraphics[width=0.49\columnwidth]{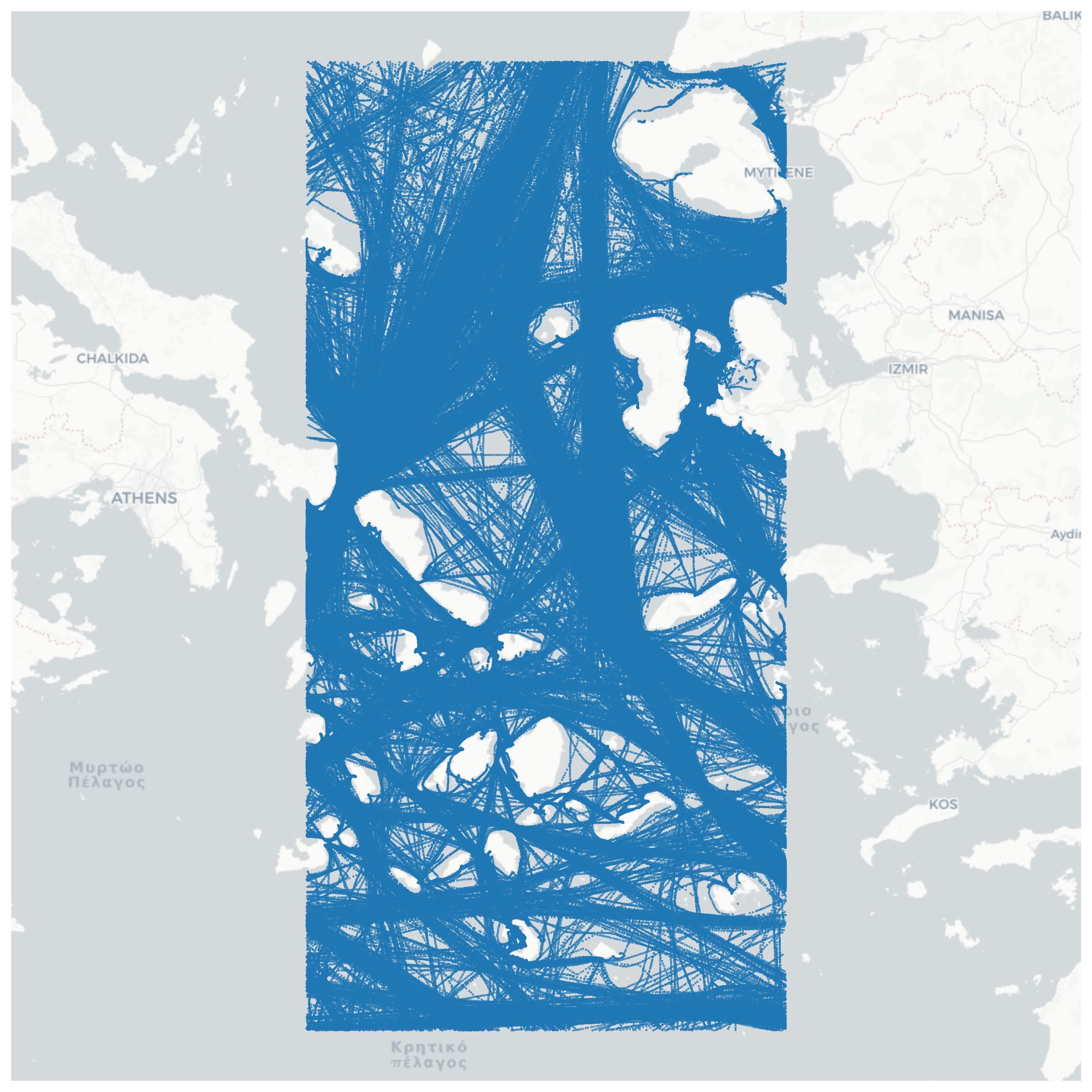}}
        \caption{Snapshot of (a) Brest; (b) Piraeus; and (c) Aegean dataset, after the preprocessing phase.}
        \label{fig:DatasetSnapshots} 
    \end{figure*}
    
    In our experimental study, we use three large-scale real-world maritime AIS datasets, referred to as ``Brest''\footnote{The dataset is publicly available at \url{https://doi.org/10.5281/zenodo.1167594}.}, ``Piraeus''\footnote{The dataset is publicly available at \url{https://doi.org/10.5281/zenodo.5562629}.}, and ``Aegean''\footnote{The (proprietary) dataset has been kindly provided by Kpler for research purposes, in the context of project MobiSpaces (\url{https://mobispaces.eu}).}, respectively. Figure \ref{fig:DatasetSnapshots} provides a visualization of these datasets on the map. All conducted experiments were implemented in Python. More specifically, the aforementioned models were implemented using PyTorch\footnote{PyTorch: An Imperative Style, High-Performance Deep Learning Library. \url{https://pytorch.org}} and trained using Flower\footnote{Flower: A Friendly Federated Learning Framework. \url{https://flower.dev}} for FL, via a GPU cluster equipped with 2 Nvidia A100 GPUs, 128 CPUs, and 1TB of RAM.

    Our data (pre-)processing stage consists of two phases. The first phase is a process consistently followed in the literature due to the noise, irregularity of sampling rate, etc. that are typical in AIS datasets. 
    The second phase (data preparation for model training) on the other hand, uses the output of the first phase as its input and is performed for ML-specific purposes, i.e., to feed its output into our (Fed)Nautilus architecture. 
    
    In particular, the preprocessing phase includes (i) record de-duplication, in which we eliminate duplicate AIS messages based on their unique ID (Maritime Mobile Service Identity - MMSI) and time of transmission, and
    (ii) dropping vessels with invalid ID with respect to the first three digits of MMSI that correspond to the country of origin (Maritime Identification Digits - MID) \cite{DBLP:journals/taes/CapobiancoMFBW21}. 
    In order to focus our model on the ``big picture'' of maritime mobility, and avoid the effect of vessels' micromovements, as an effect of the unstable sampling rate of AIS\footnote{MarineTraffic - How often do the positions of the vessels get updated on MarineTraffic?, \url{https://help.marinetraffic.com/hc/en-us/articles/217631867-How-often-do-the-positions-of-the-vessels-get-updated-on-MarineTraffic-}. Last visited at Jan. 20, 2024.}, we also subsample the vessels' trajectories, so that the minimum temporal difference between two points is set to $\Delta t\_{min} = 10$ seconds. 
    
    Because this operation may drop several -- insignificant -- AIS messages, out of the resulting trajectories we drop the ones with less than $min\_pts = 20$ points. Afterwards, we drop the AIS erroneous messages due to invalid speed (i.e., speed over $speed\_max = 50$ knots) or stationarity (i.e., speed less than $speed\_min = 1$ knot). Finally, we perform a trajectory segmentation task when a successive pair of points with a temporal difference higher than $t\_max$ is detected (keeping, also in this case, the property of at least $min\_pts$ points per trajectory). In our study, we experimented with two different thresholds, $t\_max = 30$ min. and $60$ min., in order to evaluate the prediction accuracy of our model up to $\Delta t_{\text{next}} = 30$ min. and $\Delta t_{\text{next}} = 60$ min., respectively. Table \ref{tab:dataset-stats} illustrates the statistics of the datasets (spatial and temporal range, number of locations, trajectories, etc.) after the above mentioned steps.

    Towards training the model in order to account the trajectories' variable length as well as their temporal dependencies, we take the first order difference (slope) of the kinematic features (i.e., location, speed, course) for each trajectory, and then apply a semi-overlapping sliding window with length (i.e., number of transitions) from $length\_min = 18$ up to $length\_max = 32$ transitions. Since the aim is to predict the next location, we select as label the next transition from each window. 
    Finally, to account for the unit difference in the time series variables, we normalize their corresponding values by subtracting the mean of each feature and dividing by its variance (z-score standardization).

    In the two sections that follow, we provide the experimental results of our study, comparing the two variants, centralized and federated, of the proposed (Fed)Nautilus framework presented in Section \ref{sec:Background_and_Definitions}, also with respect to the current state-of-the-art \cite{DBLP:journals/tits/ChondrodimaPPT23}. The performance comparison is based on a popular prediction quality measure, namely the Final Displacement Error (FDE) defined in Eq. \ref{eq:FDE}.

    \begin{equation}
        FDE = \displaystyle\dfrac{1}{n} \displaystyle\sum_{i} \sqrt{
            \begin{aligned}
             \left(\Delta x_i^{pred} - \Delta x_i^{true}\right)^2 + \\
             \left(\Delta y_i^{pred} - \Delta y_i^{true}\right)^2
            \end{aligned}
          }
        \label{eq:FDE}
    \end{equation}

    \subsection{Experimental Results on Nautilus}
        In this section, we demonstrate the results of our experimental study on Nautilus.
        Following the preprocessing outlined in Section \ref{subsect:ExperimentalSetup}, we acquire 265,525, 109,906, and 70,201 sliding windows from the Brest, Piraeus, and Aegean datasets, respectively. These windows will be used to train the model. They are divided into training, validation, and test sets using a split ratio of 50:25:25\% (1 fold) with respect to the datasets' temporal span (c.f., Table \ref{tab:dataset-stats}).
        After training our VLF instances, Table \ref{tab:fedvlf-cml-perf} illustrates the performance of Nautilus on the datasets' test set with respect to FDE. We observe that our solution outperforms the state-of-the-art method \cite{DBLP:journals/tits/ChondrodimaPPT23} in the majority of cases; in particular, it is the clear winner on the Brest and Piraeus datasets, especially when both models are trained over the $t\_max = 60$ min. trajectory segmentation variant of datasets, while it appears to be a balance between the two models on the Aegean dataset.

        \input{tabs/fedvlf-cml-perf}

        \begin{figure}[!ht]
        \centering
        \subfloat[\label{subfig:fl-learning-curve-brest}]{%
            \includegraphics[width=0.5\columnwidth]{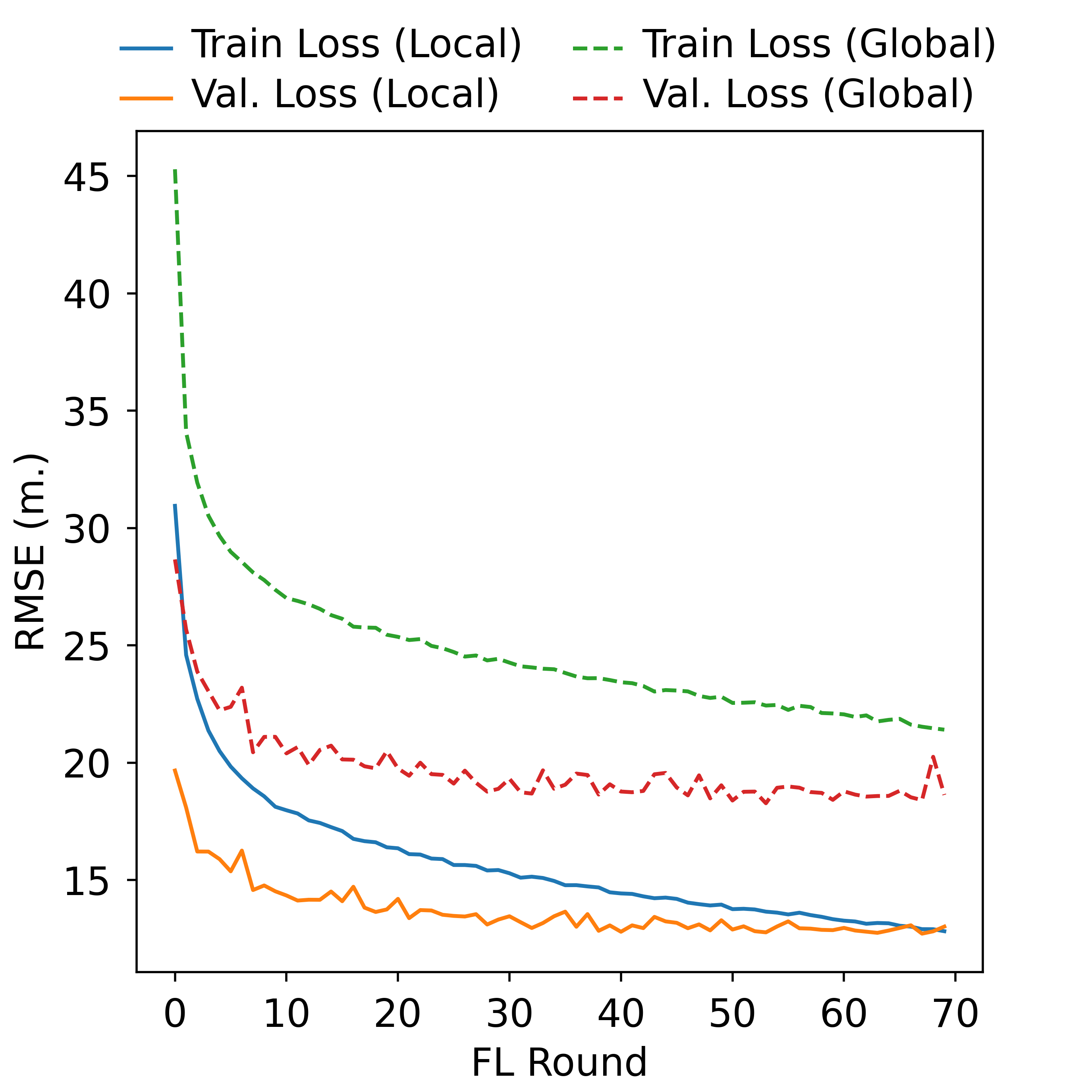}}
        \\
        \subfloat[\label{subfig:fl-learning-curve-piraeus}]{%
            \includegraphics[width=0.5\columnwidth]{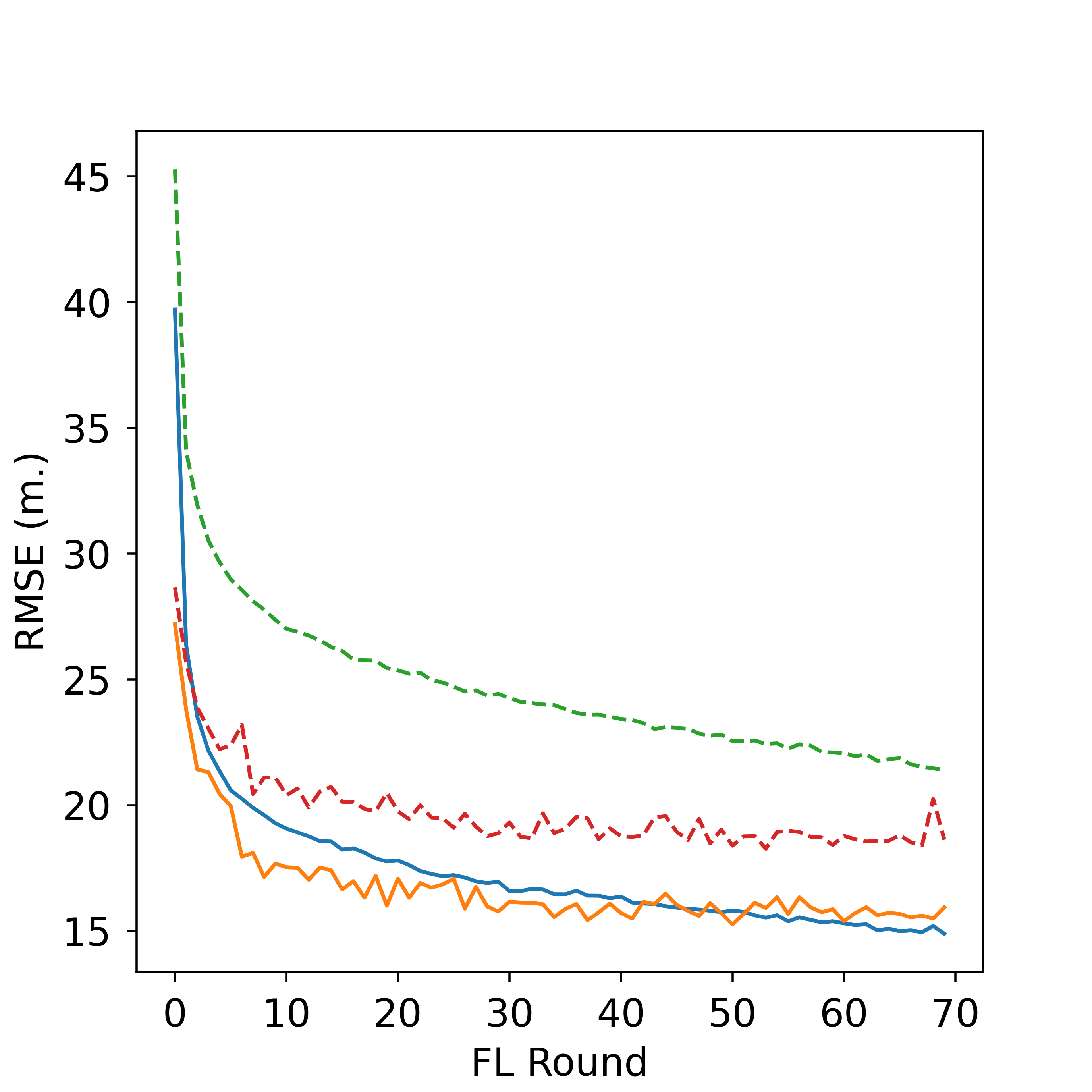}}
        \hfill
        \subfloat[\label{subfig:fl-learning-curve-marinetraffic}]{%
            \includegraphics[width=0.5\columnwidth]{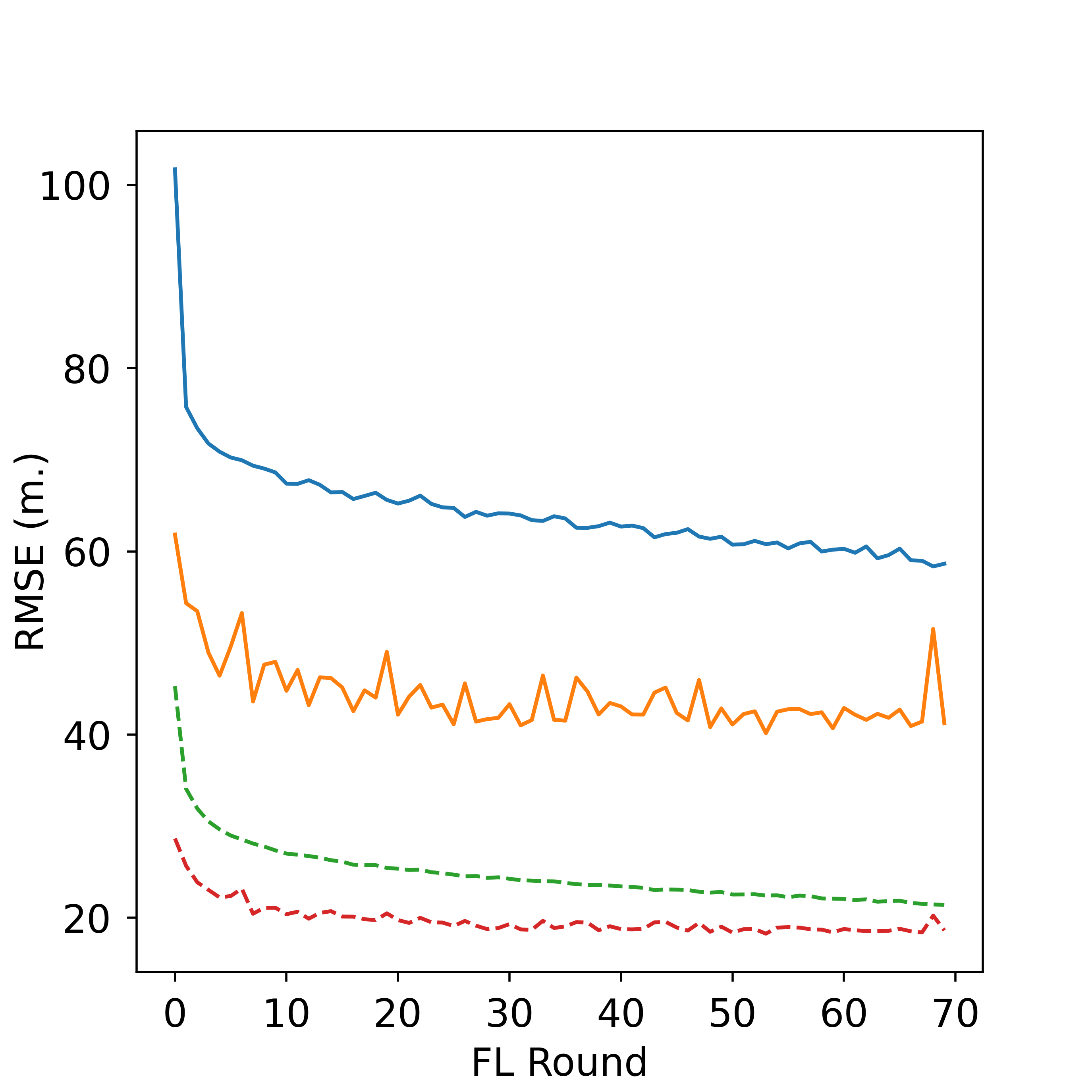}}
        \caption{Learning curve for (a) Brest, (b) Piraeus, and (c) Aegean local FedNautilus instances compared to the global FedNautilus model. Solid blue and orange lines correspond to participants’ training and validation loss, respectively, while dashed green and red lines correspond to the global FedNautilus training and validation loss, respectively.}
        \label{fig:fl-learning-curve}
        \end{figure}
        
        Focusing on the results of Nautilus, and comparing the two variants (i.e., 30 min. vs. 60 min. trajectory segmentation threshold), we observe that the 30 min. variant yields slightly better results on the Brest dataset, while the opposite happens on the Piraeus and Aegean datasets. A first conclusion out of this is that the segmentation threshold is not a critical factor for the quality of the prediction; the original assumption that by setting $t\_max = 30$ min. at the preprocessing phase would lead to consistently better performance for predictions up to $\Delta t_{\text{next}} = 30$ min. was not confirmed in practice. This behaviour can partially be attributed to the increase in the population of the train set, in terms of target lookaheads, which may introduce a regularization effect to the model.

    \subsection{Experimental Results on FedNautilus}
        Moving to the FL setting, let us assume that the participating partners are in agreement to exchange the parameters of each partner's VLF model (i.e., local model), and aggregate them into a single entity (i.e., global model) using the FedProx algorithm \cite{DBLP:conf/mlsys/LiSZSTS20} with $\mu_{prox} = 10^{-3}$, as presented in Section \ref{subsect:Methodology}. After 70 rounds, Table \ref{tab:fedvlf-fl-perf} (rows titled "FedNautilus") illustrates the performance of the (global) FedNautilus model over the test set of the datasets at hand. Compared to the prediction error of our centralized solution (c.f. Table \ref{tab:fedvlf-cml-perf}), we observe that the prediction quality on all three datasets of the global FedNautilus model has been decreased with respect to the centralized results.
        
        In order to further understand the reason behind this behaviour, Figure \ref{fig:fl-learning-curve} illustrates the learning curve of the partners' local VLF instances, compared to the learning curve of the global FedNautilus model. In general, we observe that the training/validation loss of the Aegean (c.f., Figure \ref{subfig:fl-learning-curve-marinetraffic}) local models diverges from the global FedNautilus model, a behaviour which is observed throughout the training process, better known as ``client-drift''.

        The main cause behind ``client-drift'' lies within the participating parties' heterogeneity. In particular, Figure \ref{fig:DatasetPDFs} illustrates the Kernel Density Estimation (KDE) of the first and second principal components of the three datasets. We observe that the estimated distributions of Brest and Piraeus datasets follow a - seemingly - unimodal distribution, while on the other hand, the multimodal distribution of the Aegean dataset, introduces a high level of heterogeneity, which inhibits the training process of global FedNautilus model from matching the participants' performance target(s).

        \begin{figure}[ht]
          \centering
          \subfloat[\label{subfig:BrestDatasetPDF}]{
              \includegraphics[width=0.5\columnwidth]{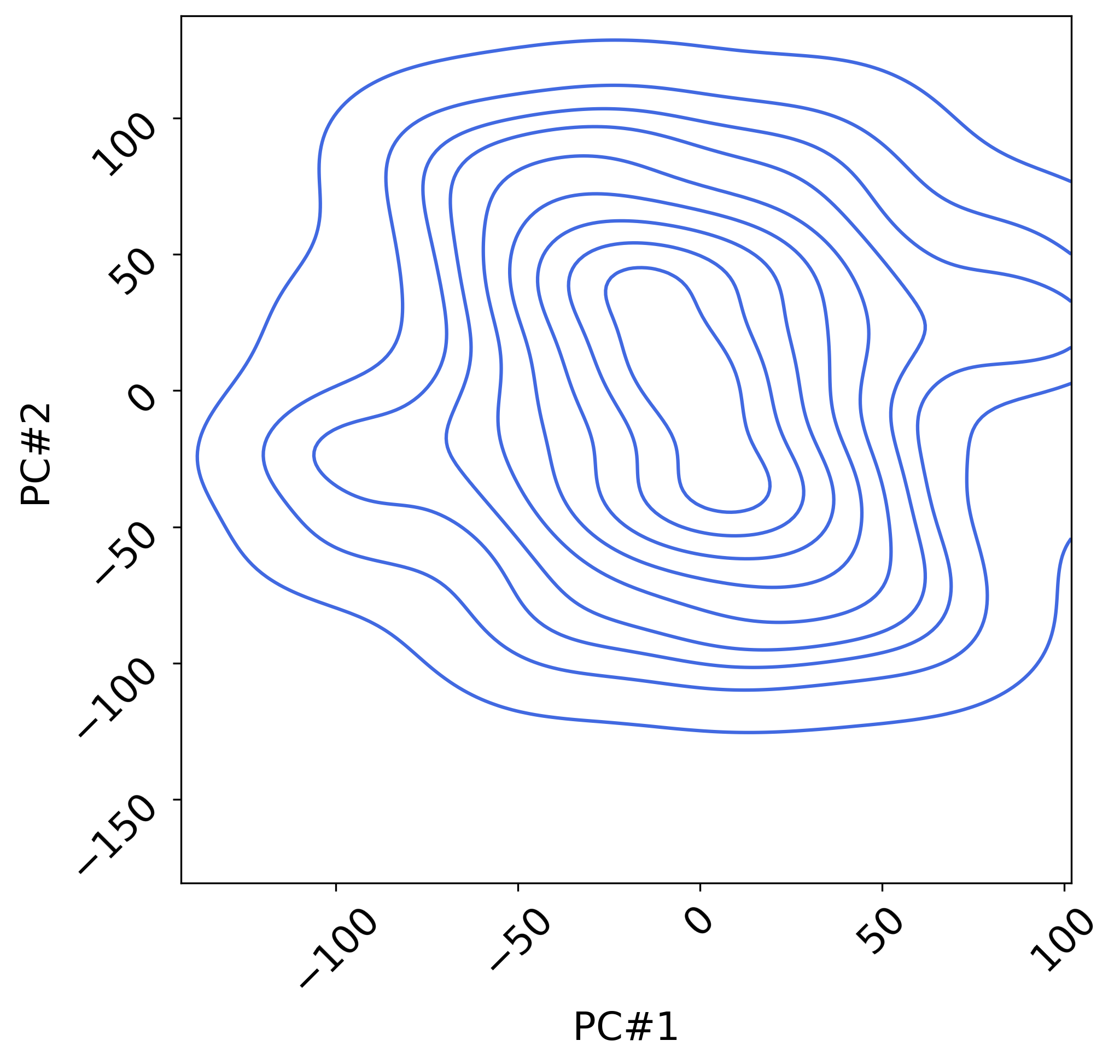}}
          \\
          \subfloat[\label{subfig:PiraeusDatasetPDF}]{%
              \includegraphics[width=0.5\columnwidth]{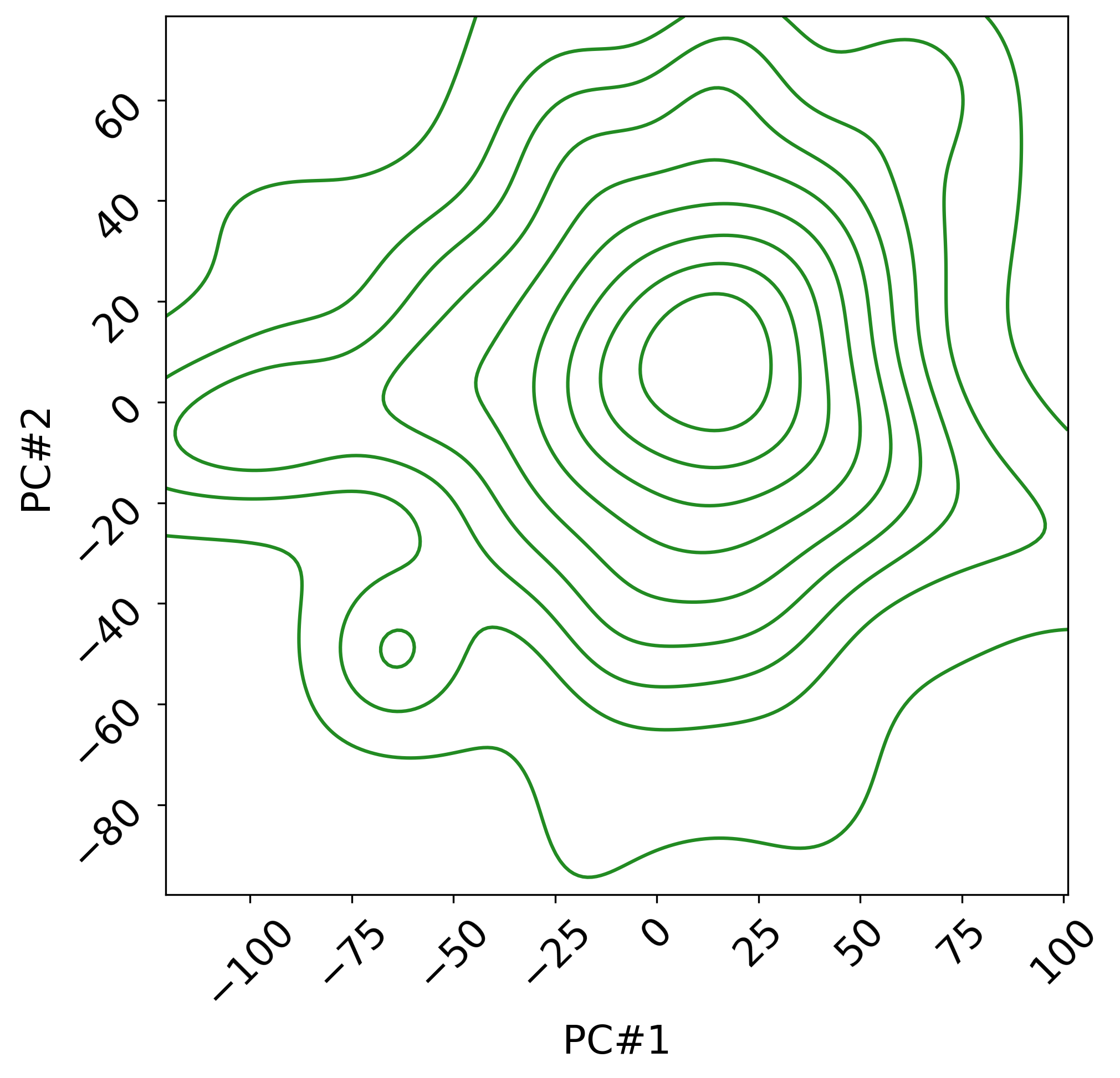}}
          \subfloat[\label{subfig:MarineTrafficDatasetPDF}]{%
              \includegraphics[width=0.5\columnwidth]{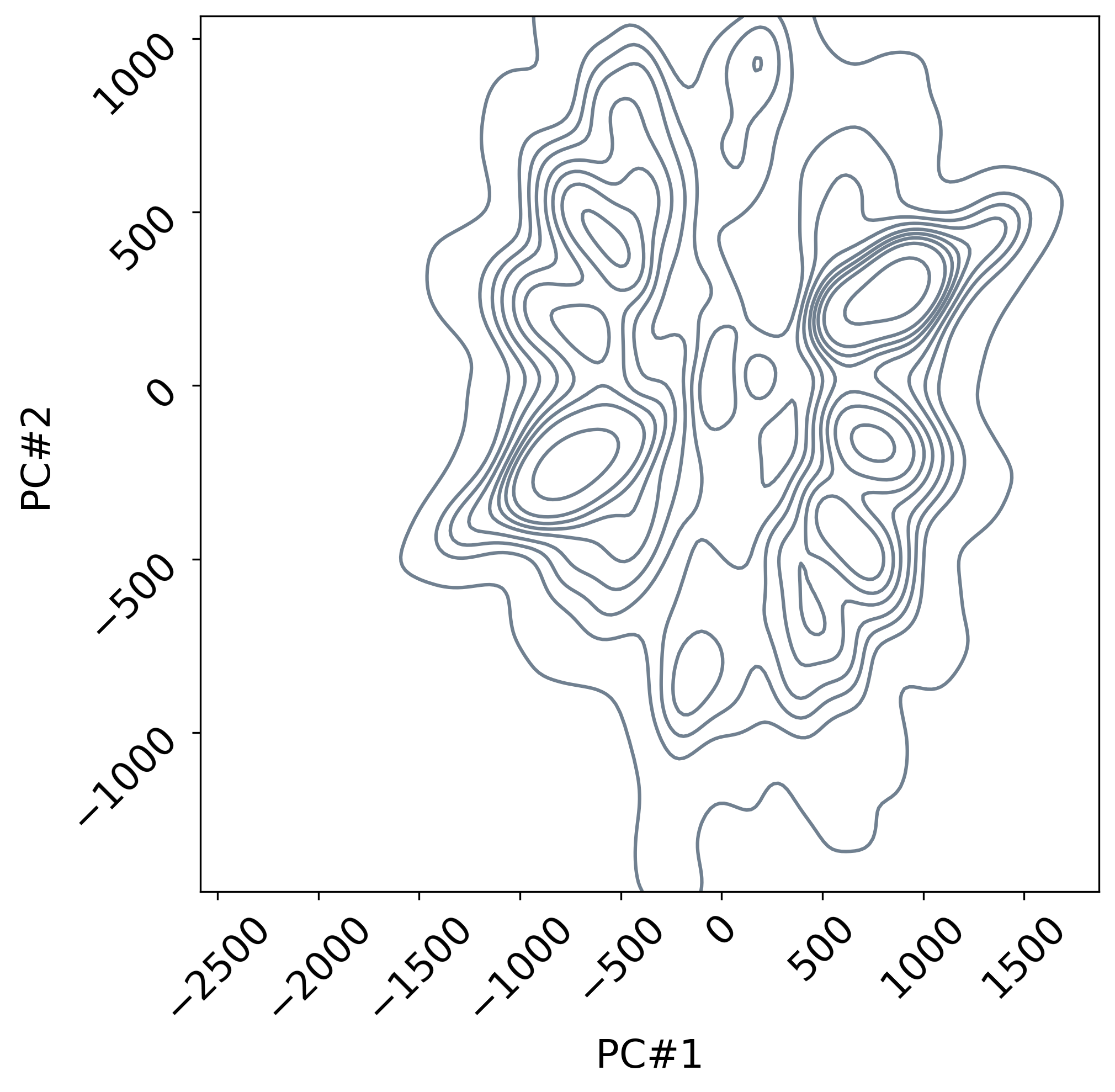}}
          \caption{Kernel Density Estimation (KDE) of 1\textsuperscript{st} and 2\textsuperscript{nd} principal components of (a) Brest; (b) Piraeus; and (c) Aegean dataset.}
          \label{fig:DatasetPDFs} 
        \end{figure}

        \input{tabs/fedvlf-fl-perf}

        By adjusting the $\mu_{prox}$ parameter, we can be either more ``strict'' or more ``relaxed'' with participating data silos whose contributions (i.e., model weights) greatly deviate with respect to the aggregated (global) model. Table \ref{tab:fedvlf-muprox-perf} illustrates the performance of (global)FedNautilus for $\mu_{prox} = \lbrace 10^{-4}, 10^{-3}, 10^{-2}, 10^{-1}, 1 \rbrace$. 
        
        Combined with the insight given by Figure \ref{fig:DatasetPDFs}, we observe that the Piraeus dataset has the most increase in performance, presenting its best at $\mu_{prox} = 1.0$, with $\mu_{prox} = 10^{-3}$ being a close contender. On the other hand, we observe that the Brest and Aegean datasets present an overall increase in FDE up to 25 min., presenting their best at $\mu_{prox} = 10^{-1}$ and $\mu_{prox} = 10^{-3}$, respectively. In other words, the more we increase $\mu_{prox}$, the more FedProx tends to benefit the Piraeus dataset, and ignore the Brest and Aegean datasets, due to their increasing complexity in terms of KDE.
        
        Focusing on the FedNautilus instances trained with $\mu_{prox} = 10^{-3}$ (more or less, the same observations hold the other values of $\mu_{prox}$ used in our experimental study), Table \ref{tab:fedvlf-fl-perf} (rows titled ``(Per)FedNautilus'') illustrates the performance of the global personalized FedNautilus instance, (Per)FedNautilus, on the three datasets. Compared to the centralized and the ``global'' federated approaches (rows titled ``Nautilus'' and ``(global)FedNautilus'', respectively, in Table \ref{tab:fedvlf-fl-perf}), it appears that personalization clearly improved the
        prediction accuracy of our model, not only against the global FL model but also against the (local) models trained upon the partners' datasets. This is a key result of our performance study that confirms the value of the FL paradigm.

    \subsection{A note on the communication cost}\label{subsec:FL-Bandwidth-Gain}
        As mentioned in Section \ref{sect:Introduction}, FL offers significant improvements in communication costs compared to centralized ML, especially when dealing with large datasets. Assume a hypothetical scenario in which, rather than training an ML-based model in a centralized (isolated) manner, each data silo pools their own datasets and trains a centralized (unified) ML-based model. The combined size of the Brest, Piraeus, and Aegean datasets is up to 4.15 GB. The server sends the updated model (6.7 MB, of which 2.2 MB is the size of the model parameters) to participating data silos at each epoch, for a total of 6.7 * 3 * 70 = 1.37 GB for 70 epochs in the worst-case scenario (i.e., the early stopping mechanism is not activated). As a result, the communication cost of the centralized training process is 5.52 GB.

        \input{tabs/fedvlf-muprox-perf}

        On the other hand, FL trains the model locally on the devices or servers where the data is stored. In this case, only the updated model parameters, 2.2 MB in total, must be sent from each data silo to the central server and vice versa. As a result, the total communication cost for each training round in FL is 6.6 MB (3 silos * 2.2 MB) * 2 (send/receive), or 0.90 GB for 70 FL rounds. This represents a significant reduction in communication cost of 84\% compared to the centralized approach. As a result, FedNautilus can be considered a viable solution for not only data silos but also edge devices with limited bandwidth resources.

\section{Conclusion}\label{sec:conclusion}  
    In this paper, we proposed (Fed)Nautilus, a VLF framework extending current state-of-the-art \cite{DBLP:journals/tits/ChondrodimaPPT23}, we trained it in two variants, following the centralized ML and the FL paradigms, respectively, and assessed the pros and cons of each approach. In particular, through an extensive experimental study on three real-world AIS datasets, we demonstrated the efficiency of Nautilus in comparison with related work. We also evaluated the benefits and open problems of the FL-based solution, FedNautilus, as well as the usefulness of personalization over heterogeneous datasets. 
    
    Following the research guidelines of the emerging Mobility Data Science era \cite{10.1145/3652158}, in the near future, we aim to further adjust the architecture of the FedNautilus model by incorporating additional external factors, such as weather conditions and itinerary information. In a parallel line of research, we aim to transfer the FedNautilus model from the cross-silo to the cross-device paradigm in order to assess the balance between prediction quality and communication cost, based on the preliminary findings about the latter, which were discussed in Section \ref{subsec:FL-Bandwidth-Gain}. In the long-term, we aim to address the client-drift issue of FedNautilus by either fine-tuning the existing algorithms, or implementing newer ones, such as \cite{DBLP:conf/nips/0001MO20}.

\section*{Acknowledgment}\addcontentsline{toc}{section}{Acknowledgment}
    \noindent
    This work was supported in part by the Horizon Framework Programme of the European Union under grant agreement No. 101070279 (MobiSpaces; \url{https://mobispaces.eu}). In this work, Kpler provided the Aegean AIS dataset and the requirements of the business case.

\bibliographystyle{IEEEtran.bst}
\bibliography{bibliography.bib}

\end{document}

%% file: tabs/dataset-stats.tex
\begin{table*}[!ht]
    \caption{Statistics of the datasets used in our experimental study, after the preprocessing phase (with two different trajectory segmentation thresholds, $t\_max = 30$ min. and $60$ min., respectively).}
    \label{tab:dataset-stats}
    \centering    
    \renewcommand{\arraystretch}{2.5}
    \resizebox{\textwidth}{!}{%
    \begin{tabular}{@{}lllllllll@{}}
    \toprule
                  & Spatial Range                                                                               & \renewcommand{\arraystretch}{1.2}\begin{tabular}[c]{@{}l@{}}Temporal \\ Range\end{tabular}          & \#Records & \renewcommand{\arraystretch}{1.2}\begin{tabular}[c]{@{}l@{}}\#Vessels \\ (i.e., distinct IDs)\end{tabular} & \#Trajectories & \renewcommand{\arraystretch}{1.2}\begin{tabular}[c]{@{}l@{}}Sampling Rate (sec.)\\ (min.; avg.; max)\end{tabular} & \renewcommand{\arraystretch}{1.2}\begin{tabular}[c]{@{}l@{}}\#Points per Trajectory\\ (min.; avg.; max.)\end{tabular} \\ \midrule
        Brest     & \renewcommand{\arraystretch}{1.2}\begin{tabular}[c]{@{}l@{}}lon $\in [-9.71, -1.09]$\\ lat $\in [45.00, 50.24]$\end{tabular} & \renewcommand{\arraystretch}{1.2}\begin{tabular}[c]{@{}l@{}}2015-10-01 – \\ 2016-03-31\end{tabular} & \renewcommand{\arraystretch}{1.2}\begin{tabular}[c]{@{}l@{}} 4,408,217 (30 min.) \\ 4,504,533 (60 min.) \end{tabular} & \renewcommand{\arraystretch}{1.2}\begin{tabular}[c]{@{}l@{}} 1,654 (30 min.) \\ 1,895 (60 min.) \end{tabular}     &   \renewcommand{\arraystretch}{1.2}\begin{tabular}[c]{@{}l@{}} 14,418 (30 min.) \\ 13,155 (60 min.) \end{tabular}    & \renewcommand{\arraystretch}{1.2}\begin{tabular}[c]{@{}l@{}} 10; 26; 1,800 (30 min.)  \\ 10; 36; 3,600 (60 min.) \end{tabular}    & \renewcommand{\arraystretch}{1.2}\begin{tabular}[c]{@{}l@{}} 20; 306; 13,638 (30 min.)  \\ 20; 342; 15,557 (60 min.) \end{tabular}         \\
        Piraeus   & \renewcommand{\arraystretch}{1.2}\begin{tabular}[c]{@{}l@{}}lon $\in [23.02, 23.80]$\\ lat $\in [37.50, 38.04]$\end{tabular} & \renewcommand{\arraystretch}{1.2}\begin{tabular}[c]{@{}l@{}}2019-01-01 – \\ 2019-03-30\end{tabular} & \renewcommand{\arraystretch}{1.2}\begin{tabular}[c]{@{}l@{}} 1,903,582 (30 min.) \\ 1,951,356 (60 min.) \end{tabular} & \renewcommand{\arraystretch}{1.2}\begin{tabular}[c]{@{}l@{}} 1,321 (30 min.) \\ 1,324 (60 min.) \end{tabular}     &   \renewcommand{\arraystretch}{1.2}\begin{tabular}[c]{@{}l@{}} 12,895 (30 min.) \\ 10,588 (60 min.) \end{tabular}    & \renewcommand{\arraystretch}{1.2}\begin{tabular}[c]{@{}l@{}} 10; 35; 1,800 (30 min.)  \\ 10; 48; 3,600 (60 min.) \end{tabular}    & \renewcommand{\arraystretch}{1.2}\begin{tabular}[c]{@{}l@{}} 20; 148; 5,659 (30 min.)  \\ 20; 184; 5,659 (60 min.) \end{tabular}          \\
        Aegean    & \renewcommand{\arraystretch}{1.2}\begin{tabular}[c]{@{}l@{}}lon $\in [24.46, 26.59]$\\ lat $\in [36.08, 39.49]$\end{tabular} & \renewcommand{\arraystretch}{1.2}\begin{tabular}[c]{@{}l@{}}2018-11-01 – \\ 2018-11-30\end{tabular} & \renewcommand{\arraystretch}{1.2}\begin{tabular}[c]{@{}l@{}} 1,216,691 (30 min.) \\ 1,226,136 (60 min.)  \end{tabular} & \renewcommand{\arraystretch}{1.2}\begin{tabular}[c]{@{}l@{}} 2,901 (30 min.) \\ 2,907 (60 min.) \end{tabular}     &   \renewcommand{\arraystretch}{1.2}\begin{tabular}[c]{@{}l@{}} 8,143 (30 min.)  \\ 7,741 (60 min.)  \end{tabular}    & \renewcommand{\arraystretch}{1.2}\begin{tabular}[c]{@{}l@{}} 11; 165; 1,800 (30 min.) \\ 11; 169; 3,600 (60 min.) \end{tabular}   & \renewcommand{\arraystretch}{1.2}\begin{tabular}[c]{@{}l@{}} 20; 149; 1,302 (30 min.)  \\ 20; 158; 1,600 (60 min.) \end{tabular}          \\ \bottomrule
    \end{tabular}%
    }
\end{table*}

%% file: tabs/fedvlf-cml-perf.tex
\begin{table*}[ht]
    \caption{Prediction error (FDE; meters) of Nautilus vs. state-of-the-art per dataset (less is better).}
    \label{tab:fedvlf-cml-perf}
    \centering    
    \renewcommand{\arraystretch}{1.3}
    \resizebox{\textwidth}{!}{%
    \begin{tabular}{llrrrrrrrrrrrr}
    \toprule
                                                &                                                         & (0,5]        & (5, 10]       & (10, 15]     & (15, 20]     & (20, 25]     & (25, 30]      & (30, 35]     & (35, 40]      & (40, 45]      & (45, 50]      & (50, 55]      & (55, 60]      \\ \midrule
    \multirow{4}{*}{\rotatebox{90}{Brest}}      & VLF-LSTM \cite{DBLP:journals/tits/ChondrodimaPPT23} (30 min.) & 13           & 308           & 602          & \textbf{802} & \textbf{1073}& 1316          & N/A          & N/A           & N/A           & N/A           & N/A           & N/A           \\ 
                                                & Nautilus (30 min.)                                      & \textbf{12}  & \textbf{302}  & \textbf{579} & 811          & 1139         & \textbf{1220} & N/A          & N/A           & N/A           & N/A           & N/A           & N/A           \\ \cmidrule{2-14}
                                                & VLF-LSTM \cite{DBLP:journals/tits/ChondrodimaPPT23} (60 min.) & 14           & 366           & 781          & 1042         & 1210         & 1427          & 2926         & 2493          & 2699          & 1987          & 3991          & \textbf{4167} \\ 
                                                & Nautilus (60 min.)                                      & \textbf{13}  & \textbf{353}  & \textbf{714} & \textbf{893} & \textbf{1032}& \textbf{1380} & \textbf{2654}& \textbf{2296} & \textbf{2119} & \textbf{1897} & \textbf{3627} & 4177          \\ \midrule
    \multirow{4}{*}{\rotatebox{90}{Piraeus}}    & VLF-LSTM \cite{DBLP:journals/tits/ChondrodimaPPT23} (30 min.) & 14           & 157           & \textbf{348} & \textbf{438} & 713          & \textbf{551}  & N/A          & N/A           & N/A           & N/A           & N/A           & N/A           \\ 
                                                & Nautilus (30 min.)                                      & \textbf{14}  & \textbf{154}  & 353          & 442          & \textbf{685} & 601           & N/A          & N/A           & N/A           & N/A           & N/A           & N/A           \\ \cmidrule{2-14}
                                                & VLF-LSTM \cite{DBLP:journals/tits/ChondrodimaPPT23} (60 min.) & 15           & 155           & \textbf{332} & 436          & 570          & 406           & 672          & 478           & 871           & 1458          & 1518          & 2620          \\ 
                                                & Nautilus (60 min.)                                      & \textbf{15}  & \textbf{144}  & 334          & \textbf{409} & \textbf{537} & \textbf{332}  & \textbf{639} & \textbf{353}  & \textbf{723}  & \textbf{1369} & \textbf{1346} & \textbf{2179} \\ \midrule
    \multirow{4}{*}{\rotatebox{90}{Aegean}}     & VLF-LSTM \cite{DBLP:journals/tits/ChondrodimaPPT23} (30 min.) & \textbf{41}  & 201           & \textbf{477} & 505          & 1036         & \textbf{1764} & N/A          & N/A           & N/A           & N/A           & N/A           & N/A           \\ 
                                                & Nautilus (30 min.)                                      & 43           & \textbf{198}  & 494          & \textbf{501} & \textbf{902} & 1976          & N/A          & N/A           & N/A           & N/A           & N/A           & N/A           \\ \cmidrule{2-14}
                                                & VLF-LSTM \cite{DBLP:journals/tits/ChondrodimaPPT23} (60 min.) & \textbf{43}  & 190           & 499          & \textbf{348} & \textbf{1010}& 1302          & 967          & \textbf{2360} & 1460          & 2950          & \textbf{8369} & 5124          \\ 
                                                & Nautilus (60 min.)                                      & 45           & \textbf{188}  & \textbf{482} & 457          & 1053         & \textbf{1132} & \textbf{918} & 2684          & \textbf{1373} & \textbf{2860} & 9402          & \textbf{2848} \\  \bottomrule
    \end{tabular}
    }
\end{table*}

%% file: tabs/fedvlf-fl-perf.tex
\begin{table*}[!t]
    \caption{Prediction error (FDE; meters) of Nautilus vs. FedNautilus per dataset (less is better).}
    \label{tab:fedvlf-fl-perf}
    \centering    
    \renewcommand{\arraystretch}{1.3}
    \resizebox{\textwidth}{!}{%
    \begin{tabular}{llrrrrrrrrrrrr}
    \toprule
                                                            &                                   & (0,5]        & (5, 10]       & (10, 15]     & (15, 20]     & (20, 25]     & (25, 30]      & (30, 35]      & (35, 40]      & (40, 45]         & (45, 50]      & (50, 55]      & (55, 60]      \\ \midrule
\multirow{7}{*}{\rotatebox{90}{Brest}}                      & Nautilus (30 min.)                & 12           & 302           & 579          & 811          & 1139         & 1220          & N/A           & N/A           & N/A              & N/A           & N/A           & N/A           \\
                                                            & (global)FedNautilus (30 min.)     & 14           & 332           & 667          & 968          & 1295         & 1577          & N/A           & N/A           & N/A              & N/A           & N/A           & N/A           \\
                                                            & (Per)FedNautilus (30 min.)        & \textbf{12}  & \textbf{291}  & \textbf{547} & \textbf{802} & \textbf{1157}& \textbf{1170} & N/A           & N/A           & N/A              & N/A           & N/A           & N/A           \\ \cmidrule{2-14}
                                                            & Nautilus (60 min.)                & 13           & 353           & 714          & 893          & 1032         & 1380          & 2654          & \textbf{2296} & 2119             & \textbf{1897} & 3627          & \textbf{4177} \\ 
                                                            & (global)FedNautilus (60 min.)     & 17           & 375           & 729          & 1033         & 1058         & 1579          & 2715          & 2480          & 2285             & 2222          & \textbf{3296} & 4308          \\ 
                                                            & (Per)FedNautilus (60 min.)        & 13           & \textbf{345}  & \textbf{681} & \textbf{864} & \textbf{907} & \textbf{1250} & \textbf{2569} & 2411          & \textbf{1926}    & 2040          & 3367          & 4215          \\ \midrule
\multirow{7}{*}{\rotatebox{90}{Piraeus}}                    & Nautilus (30 min.)                & 14           & 154           & 353          & 442          & 685          & 601           & N/A           & N/A           & N/A              & N/A           & N/A           & N/A           \\
                                                            & (global)FedNautilus (30 min.)     & 23           & 226           & 409          & 520          & 775          & 728           & N/A           & N/A           & N/A              & N/A           & N/A           & N/A           \\
                                                            & (Per)FedNautilus (30 min.)        & \textbf{13}  & \textbf{140}  & \textbf{308} & \textbf{356} & \textbf{607} & \textbf{568}  & N/A           & N/A           & N/A              & N/A           & N/A           & N/A           \\ \cmidrule{2-14}
                                                            & Nautilus (60 min.)                & 15           & 144           & 334          & 409          & 537          & \textbf{332}  & 639           & \textbf{353}  & \textbf{723}     & \textbf{1369} & \textbf{1346} & 2179          \\ 
                                                            & (global)FedNautilus (60 min.)     & 24           & 239           & 387          & 534          & 654          & 675           & 867           & 772           & 1238             & 1555          & 1708          & 2767          \\ 
                                                            & (Per)FedNautilus (60 min.)        & \textbf{15}  & \textbf{144}  & \textbf{298} & \textbf{363} & \textbf{537} & 412           & \textbf{636}  & 542           & 914              & 1462          & 1347          & \textbf{2007} \\ \midrule
\multirow{7}{*}{\rotatebox{90}{Aegean}}                     & Nautilus (30 min.)                & 43           & \textbf{198}  & \textbf{494} & 501          & \textbf{902} & 1976          & N/A           & N/A           & N/A              & N/A           & N/A           & N/A           \\
                                                            & (global)FedNautilus (30 min.)     & 550          & 1810          & 3747         & 4035         & 5175         & 9096          & N/A           & N/A           & N/A              & N/A           & N/A           & N/A           \\
                                                            & (Per)FedNautilus (30 min.)        & \textbf{43}  & 211           & 541          & \textbf{482} & 1177         & \textbf{1608} & N/A           & N/A           & N/A              & N/A           & N/A           & N/A           \\ \cmidrule{2-14}
                                                            & Nautilus (60 min.)                & \textbf{45} & \textbf{188}  & 482          & 457          & 1053          & \textbf{1132} & \textbf{918}  & \textbf{2684} & 1373             & \textbf{2860} & \textbf{9402} & \textbf{2848} \\ 
                                                            & (global)FedNautilus (60 min.)     & 528         & 1334          & 2976         & 3293         & 3109          & 8808          & 9400          & 4296          & 3226             & 7585          & 18786         & 21302         \\ 
                                                            & (Per)FedNautilus (60 min.)        & 45          & 203           & \textbf{465} & \textbf{358} & \textbf{886}  & \textbf{914}  & 1110          & 2981          & \textbf{1236}    & 2882          & 10015          & 4391          \\ \bottomrule
    \end{tabular}
    }
\end{table*}

%% file: tabs/fedvlf-muprox-perf.tex
\begin{table*}[!ht]
    \caption{Prediction error (FDE; meters) of (global)FedNautilus across different values of $\mu_{prox}$ (less is better).}
    \label{tab:fedvlf-muprox-perf}
    \centering    
    \renewcommand{\arraystretch}{1.3}
    \resizebox{\textwidth}{!}{%
    \begin{tabular}{@{}lrrrrrrrrrrrrr@{}}
    \toprule
                                             &                                  & (0,5]           & (5, 10]          & (10, 15]          & (15, 20]          & (20, 25]          & (25, 30]          & (30, 35]        & (35, 40]         & (40, 45]          & (45, 50]          & (50, 55]      & (55, 60]      \\ \midrule
    \multirow{5}{*}{\rotatebox{90}{Brest}}   & $\mu_{prox} = 0.0001$ & 17              & 393              & 754               & 1079              & 1168              & 1574              & 2755            & 2524             & 2307              & \textbf{2146}     & 3549          & 4289          \\
                                             & $\mu_{prox} = 0.001$  & 17              & \textbf{375}     & 729               & \textbf{1033}     & 1058              & 1579              & 2715            & 2480             & 2285              & 2222              & \textbf{3296} & 4308          \\
                                             & $\mu_{prox} = 0.01$   & 17              & 383              & 787               & 1059              & 1083              & 1584              & 2741            & \textbf{2430}    & 2309              & 2206              & 3464          & \textbf{4187} \\
                                             & $\mu_{prox} = 0.1$    & \textbf{17}     & 376              & \textbf{728}      & 1035              & \textbf{1037}     & \textbf{1556}     & \textbf{2682}   & 2437             & \textbf{2156}     & 2167              & 3381          & 4206          \\
                                             & $\mu_{prox} = 1.0$    & 17              & 391              & 800               & 1070              & 1153              & 1660              & 2831            & 2502             & 2329              & 2556              & 3401          & 4216          \\ \midrule
    \multirow{5}{*}{\rotatebox{90}{Piraeus}} & $\mu_{prox} = 0.0001$ & 25              & 258              & 390               & 560               & 666               & 676               & 812             & 844              & 1220              & 1634              & 1547          & 2766          \\
                                             & $\mu_{prox} = 0.001$  & 24              & 239              & 387               & 534               & 654               & 675               & 867             & 772              & 1238              & \textbf{1555}     & 1708          & 2767          \\
                                             & $\mu_{prox} = 0.01$   & \textbf{23}     & 236              & 386               & 529               & 653               & 586               & 777             & \textbf{718}     & 1140              & 1678              & 1503          & \textbf{2395} \\
                                             & $\mu_{prox} = 0.1$    & 24              & 249              & 405               & 552               & 661               & 593               & \textbf{760}    & 770              & 1184              & 1614              & 1584          & 2531          \\
                                             & $\mu_{prox} = 1.0$    & 24              & \textbf{234}     & \textbf{365}      & \textbf{504}      & \textbf{626}      & \textbf{583}      & 791             & 656              & \textbf{1024}     & 1568              & \textbf{1488} & 2592          \\ \midrule
    \multirow{5}{*}{\rotatebox{90}{Aegean}}  & $\mu_{prox} = 0.0001$ & 561             & 1532             & 3519              & 3595              & 3680              & \textbf{8144}     & \textbf{8304}   & 6039             & 4390              & 7654              & 19453         & 30318         \\
                                             & $\mu_{prox} = 0.001$  & \textbf{528}    & \textbf{1334}    & \textbf{2976}     & \textbf{3293}     & \textbf{3109}     & 8808              & 9400            & \textbf{4296}    & \textbf{3236}     & 7585              & 18786         & \textbf{21302}\\
                                             & $\mu_{prox} = 0.01$   & 582             & 1440             & 3005              & 3555              & 3267              & 8158              & 9314            & 5295             & 4582              & 5997              & 16759         & 32592         \\
                                             & $\mu_{prox} = 0.1$    & 567             & 1491             & 3368              & 3743              & 4303              & 8483              & 9679            & 5815             & 3780              & 8207              & 19034         & 34912         \\
                                             & $\mu_{prox} = 1.0$    & 552             & 1425             & 3123              & 3311              & 3416              & 7924              & 8562            & 5097             & 4205              & \textbf{4870}     & \textbf{15212}& 27838         \\ \bottomrule
    \end{tabular}
    }
\end{table*}

%% file: manuscript.bbl
\begin{thebibliography}{10}
\providecommand{\url}[1]{#1}
\csname url@samestyle\endcsname
\providecommand{\newblock}{\relax}
\providecommand{\bibinfo}[2]{#2}
\providecommand{\BIBentrySTDinterwordspacing}{\spaceskip=0pt\relax}
\providecommand{\BIBentryALTinterwordstretchfactor}{4}
\providecommand{\BIBentryALTinterwordspacing}{\spaceskip=\fontdimen2\font plus
\BIBentryALTinterwordstretchfactor\fontdimen3\font minus
  \fontdimen4\font\relax}
\providecommand{\BIBforeignlanguage}[2]{{%
\expandafter\ifx\csname l@#1\endcsname\relax
\typeout{** WARNING: IEEEtran.bst: No hyphenation pattern has been}%
\typeout{** loaded for the language `#1'. Using the pattern for}%
\typeout{** the default language instead.}%
\else
\language=\csname l@#1\endcsname
\fi
#2}}
\providecommand{\BIBdecl}{\relax}
\BIBdecl

\bibitem{Tampakis2021i4sea}
P.~Tampakis, E.~Chondrodima, A.~Tritsarolis \emph{et~al.}, ``i4sea: a big data
  platform for sea area monitoring and analysis of fishing vessels activity,''
  \emph{Geo-spatial Information Science}, vol.~25, no.~2, pp. 132--154, 2022.

\bibitem{DBLP:conf/mbdw/MandalisCKPT22}
P.~Mandalis, E.~Chondrodima, Y.~Kontoulis \emph{et~al.}, ``Machine learning
  models for vessel traffic flow forecasting: An experimental comparison,'' in
  \emph{{Procedings of the 23rd IEEE International Conference on Mobile Data
  Management (MDM)}}, 2022.

\bibitem{DBLP:conf/mdm/TritsarolisCPT22}
A.~Tritsarolis, E.~Chondrodima, N.~Pelekis \emph{et~al.}, ``Vessel collision
  risk assessment using {AIS} data: {A} machine learning approach,'' in
  \emph{Proceedings of the 23rd IEEE International Conference on Mobile Data
  Management (MDM)}, 2022.

\bibitem{DBLP:conf/edbt/TritsarolisCTP21}
A.~Tritsarolis, E.~Chondrodima, P.~Tampakis \emph{et~al.}, ``Predicting
  co-movement patterns in mobility data,'' \emph{GeoInformatica}, 2022.

\bibitem{DBLP:series/synthesis/2019YangLCKCY}
Q.~Yang, Y.~Liu, Y.~Cheng \emph{et~al.}, \emph{Federated Learning}, ser.
  Synthesis Lectures on Artificial Intelligence and Machine Learning.\hskip 1em
  plus 0.5em minus 0.4em\relax Morgan {\&} Claypool Publishers, 2019.

\bibitem{DBLP:journals/access/ZissisCSV20}
D.~Zissis, K.~Chatzikokolakis, G.~Spiliopoulos \emph{et~al.}, ``A distributed
  spatial method for modeling maritime routes,'' \emph{{IEEE} Access}, vol.~8,
  pp. 47\,556--47\,568, 2020.

\bibitem{DBLP:conf/bigdataconf/SpiliopoulosCZB17}
G.~Spiliopoulos, K.~Chatzikokolakis, D.~Zissis \emph{et~al.}, ``Knowledge
  extraction from maritime spatiotemporal data: An evaluation of clustering
  algorithms on big data,'' in \emph{Proceedings of {IEEE} International
  Conference on Big Data}, 2017.

\bibitem{DBLP:journals/corr/McMahanMRA16}
H.~B. McMahan, E.~Moore, D.~Ramage \emph{et~al.}, ``Federated learning of deep
  networks using model averaging,'' \emph{CoRR}, vol. abs/1602.05629, 2016.

\bibitem{DBLP:journals/corr/KonecnyMYRSB16}
J.~Kone{\v{c}}n{\'y}, H.~B. McMahan, F.~X. Yu \emph{et~al.}, ``Federated
  learning: Strategies for improving communication efficiency,'' \emph{CoRR},
  vol. abs/1610.05492, 2016.

\bibitem{DBLP:journals/corr/abs-2102-07627}
X.~Qiu, T.~Parcollet, J.~Fern{\'{a}}ndez{-}Marqu{\'{e}}s \emph{et~al.}, ``A
  first look into the carbon footprint of federated learning,'' \emph{CoRR},
  vol. abs/2102.07627, 2021.

\bibitem{DBLP:journals/tits/ChondrodimaPPT23}
E.~Chondrodima, N.~Pelekis, A.~Pikrakis \emph{et~al.}, ``An efficient {LSTM}
  neural network-based framework for vessel location forecasting,''
  \emph{{IEEE} Transactions on Intelligent Transportation Systems}, vol.~24,
  no.~5, pp. 4872--4888, 2023.

\bibitem{DBLP:conf/ssd/PetrouNTGKSPVGC19}
P.~Petrou, P.~Nikitopoulos, P.~Tampakis \emph{et~al.}, ``{ARGO:} {A} big data
  framework for online trajectory prediction,'' in \emph{Proceedings of the
  16th International Symposium on Spatial and Temporal Databases ({SSTD})},
  2019.

\bibitem{DBLP:conf/bigdataconf/TampakisPDT19}
P.~Tampakis, N.~Pelekis, C.~Doulkeridis \emph{et~al.}, ``Scalable distributed
  subtrajectory clustering,'' in \emph{Proceedings of the {IEEE} International
  Conference on Big Data}, 2019.

\bibitem{DBLP:journals/access/ZygourasTZ24}
N.~Zygouras, A.~Troupiotis{-}Kapeliaris, and D.~Zissis, ``Envclus*: Extracting
  common pathways for effective vessel trajectory forecasting,'' \emph{{IEEE}
  Access}, vol.~12, pp. 3860--3873, 2024.

\bibitem{DBLP:journals/corr/abs-1906-04928}
S.~Wang, J.~Cao, and P.~S. Yu, ``Deep learning for spatio-temporal data mining:
  {A} survey,'' \emph{CoRR}, vol. abs/1906.04928, 2019.

\bibitem{DBLP:journals/taes/CapobiancoMFBW21}
S.~Capobianco, L.~M. Millefiori, N.~Forti \emph{et~al.}, ``Deep learning
  methods for vessel trajectory prediction based on recurrent neural
  networks,'' \emph{{IEEE} Transactions on Aerospace and Electronic Systems},
  vol.~57, no.~6, pp. 4329--4346, 2021.

\bibitem{DBLP:journals/sensors/SuoCCY20}
Y.~Suo, W.~Chen, C.~Claramunt \emph{et~al.}, ``A ship trajectory prediction
  framework based on a recurrent neural network,'' \emph{Sensors}, vol.~20,
  no.~18, 2020.

\bibitem{DBLP:conf/kdd/EsterKSX96}
M.~Ester, H.~Kriegel, J.~Sander \emph{et~al.}, ``A density-based algorithm for
  discovering clusters in large spatial databases with noise,'' in
  \emph{{Proceedings of the 2nd International Conference on Knowledge Discovery
  and Data Mining (KDD)}}, 1996.

\bibitem{DBLP:conf/icdm/Liu0SL19}
H.~Liu, H.~Wu, W.~Sun \emph{et~al.}, ``Spatio-temporal {GRU} for trajectory
  classification,'' in \emph{{Proceedings of IEEE International Conference on
  Data Mining (ICDM)}}, 2019.

\bibitem{doi:10.3138/FM57-6770-U75U-7727}
D.~H. Douglas and T.~K. Peucker, ``Algorithms for the reduction of the number
  of points required to represent a digitized line or its caricature,''
  \emph{Cartographica: The International Journal for Geographic Information and
  Geovisualization}, vol.~10, no.~2, pp. 112--122, 1973.

\bibitem{DBLP:conf/www/LiuCW17}
T.~Liu, W.~Chen, and T.~Wang, ``Distributed machine learning: Foundations,
  trends, and practices,'' in \emph{{Proceedings of the 26th International
  Conference on World Wide Web Companion (WWW)}}, 2017.

\bibitem{DBLP:conf/aistats/McMahanMRHA17}
B.~McMahan, E.~Moore, D.~Ramage \emph{et~al.}, ``Communication-efficient
  learning of deep networks from decentralized data,'' in \emph{{Proceedings of
  the 20th International Conference on Artificial Intelligence and Statistics
  (AISTATS}}, 2017.

\bibitem{DBLP:journals/ftml/KairouzMABBBBCC21}
P.~Kairouz, H.~B. McMahan, B.~Avent \emph{et~al.}, ``Advances and open problems
  in federated learning,'' \emph{Foundations and Trends in Machine Learning},
  vol.~14, no. 1-2, pp. 1--210, 2021.

\bibitem{DBLP:journals/corr/KonecnyMRR16}
J.~Kone{\v{c}}n{\'y}, H.~B. McMahan, D.~Ramage \emph{et~al.}, ``Federated
  optimization: Distributed machine learning for on-device intelligence,''
  \emph{CoRR}, vol. abs/1610.02527, 2016.

\bibitem{DBLP:conf/icml/KarimireddyKMRS20}
S.~P. Karimireddy, S.~Kale, M.~Mohri \emph{et~al.}, ``{SCAFFOLD:} stochastic
  controlled averaging for federated learning,'' in \emph{{Proceedings of the
  37th International Conference on Machine Learning (ICML)}}, vol. 119.\hskip
  1em plus 0.5em minus 0.4em\relax {PMLR}, 2020, pp. 5132--5143.

\bibitem{DBLP:conf/mlsys/LiSZSTS20}
T.~Li, A.~K. Sahu, M.~Zaheer \emph{et~al.}, ``Federated optimization in
  heterogeneous networks,'' in \emph{{Proceedings of the 3rd Conference on
  Machine Learning and Systems (MLSys)}}, 2020.

\bibitem{DBLP:conf/iclr/LiSBS20}
T.~Li, M.~Sanjabi, A.~Beirami \emph{et~al.}, ``Fair resource allocation in
  federated learning,'' in \emph{{Proceedings of the 8th International
  Conference on Learning Representations (ICLR)}}, 2020.

\bibitem{DBLP:conf/cvpr/WangCW022}
C.~Wang, X.~Chen, J.~Wang \emph{et~al.}, ``{ATPFL:} automatic trajectory
  prediction model design under federated learning framework,'' in
  \emph{{Proceedings of the Conference on Computer Vision and Pattern
  Recognition (CVPR)}}, 2022.

\bibitem{https://doi.org/10.1002/int.22987}
M.~Han, K.~Xu, S.~Ma \emph{et~al.}, ``Federated learning-based trajectory
  prediction model with privacy preserving for intelligent vehicle,''
  \emph{International Journal of Intelligent Systems}, vol.~37, no.~12, pp.
  10\,861--10\,879, 2022.

\bibitem{Hochreiter1997}
S.~Hochreiter and J.~Schmidhuber, ``Long short-term memory,'' \emph{Neural
  Computation}, vol.~9, no.~8, pp. 1735--1780, 1997.

\bibitem{58337}
P.~J. Werbos, ``Backpropagation through time: what it does and how to do it,''
  \emph{Proceedings of the IEEE}, vol.~78, no.~10, pp. 1550--1560, 1990.

\bibitem{DBLP:journals/corr/KingmaB14}
D.~P. Kingma and J.~Ba, ``Adam: {A} method for stochastic optimization,'' in
  \emph{{Proceedings of the 3rd International Conference for Learning
  Representations (ICLR)}}, 2015.

\bibitem{DBLP:series/lncs/Prechelt12}
L.~Prechelt, ``Early stopping - but when?'' in \emph{Neural Networks: Tricks of
  the Trade (2nd ed.)}, ser. Lecture Notes in Computer Science.\hskip 1em plus
  0.5em minus 0.4em\relax Springer, 2012, vol. 7700, pp. 53--67.

\bibitem{9743558}
A.~Z. Tan, H.~Yu, L.~Cui \emph{et~al.}, ``Towards personalized federated
  learning,'' \emph{IEEE Transactions on Neural Networks and Learning Systems},
  vol.~34, no.~12, pp. 9587--9603, 2023.

\bibitem{DBLP:conf/icc/TaikC20}
A.~Ta{\"{\i}}k and S.~Cherkaoui, ``Electrical load forecasting using edge
  computing and federated learning,'' in \emph{{Proceedings of the
  International Conference on Communications (ICC)}}, 2020.

\bibitem{10.1145/3652158}
M.~Mokbel, M.~Sakr, L.~Xiong \emph{et~al.}, ``Mobility data science:
  Perspectives and challenges,'' \emph{ACM Transactions on Spatial Algorithms
  and Systems}, 2024.

\bibitem{DBLP:conf/nips/0001MO20}
A.~Fallah, A.~Mokhtari, and A.~E. Ozdaglar, ``Personalized federated learning
  with theoretical guarantees: {A} model-agnostic meta-learning approach,'' in
  \emph{Proceedings of the 34th International Conference on Neural Information
  Processing Systems (NeurIPS)}, 2020.

\end{thebibliography}
